\documentclass{article}
\usepackage{iclr2027_conference,times}

\usepackage{amsmath,amsfonts,bm}

\def\eqref#1{equation~\ref{#1}}

\def\1{\bm{1}}

\DeclareMathAlphabet{\mathsfit}{\encodingdefault}{\sfdefault}{m}{sl}
\SetMathAlphabet{\mathsfit}{bold}{\encodingdefault}{\sfdefault}{bx}{n}

\usepackage{hyperref}
\hypersetup{colorlinks=true, citecolor=[rgb]{0.1,0.25,0.55}, linkcolor=[rgb]{0.1,0.25,0.55}, urlcolor=[rgb]{0.1,0.25,0.55}} %
\usepackage{url}
\usepackage[T1]{fontenc} %
\usepackage[utf8]{inputenc}
\usepackage{amsmath,amssymb}
\usepackage{graphicx}
\usepackage{booktabs}
\usepackage{microtype}
\usepackage{enumitem}
\usepackage{xcolor}
\usepackage{tikz}
\usepackage{placeins} %
\newenvironment{cdraft}{\par\color{black}}{\par}
\newcommand{\cd}[1]{{\color{black}#1}}
\providecommand{\cd}[1]{#1}
\makeatletter
\DeclareRobustCommand{\papercaption}[1]{%
  \ifcsname pcap@#1\endcsname\csname pcap@#1\endcsname
  \else{\color{red}[missing caption: #1 -- add it to figures/captions.tex]}\fi}
\newcommand{\setpapercaption}[2]{\expandafter\def\csname pcap@#1\endcsname{#2}}
\makeatother

\setpapercaption{tab:datasets}{\cd{Dataset statistics. Train, Val and Test give the number of time steps in each split.}}

\setpapercaption{tab:hyperparameters-shared}{\cd{Hyperparameters shared by all datasets, horizons and both backbones.}}

\setpapercaption{tab:hyperparameters-gridtst}{\cd{Per-dataset GridTST settings; a/b/c/d are the values for the four horizons in order.}}

\setpapercaption{tab:hyperparameters-itransformer-embedder}{\cd{Per-dataset iTransformer and embedder settings; a/b/c/d are the values for the four horizons in order; cal.: weight of the calendar term in the ranking.}}

\setpapercaption{tab:parameter-counts}{\cd{Trainable parameters in millions (ranges over horizons). \ourmethod{}: forecaster plus retrieval embedder; ratio: \ourmethod{} / backbone.}}

\setpapercaption{tab:seeds}{\cd{Test MSE, mean $\pm$ standard deviation over three seeds (2023, 2024, 2025), each seed first averaged over the four horizons. Reduction: MSE of \ourmethod{} with GridTST relative to the plain backbone (seed means). Seeds won: seeds in which \ourmethod{} is better.}}

\setpapercaption{tab:ablations}{\cd{Ablations with the GridTST backbone, horizon-averaged, single seed. Positive: worse without the component.}}

\setpapercaption{tab:oracle-ceiling}{\cd{Oracle retrieval ceiling on ETTh at $H=720$ (diagnostic; oracle rows choose windows with the true future from the same causally eligible bank). Level-free futures are shifted to start at the query's last observed value.}}

\setpapercaption{tab:retrieved-future-quality}{\cd{Quality of the retrieved futures, without any forecasting model: \% change of the MSE between retrieved slot futures and the query's true future on the test set (standardized scale, mean over the four horizons; negative = closer to the truth). Mean of slots: the average of the $K$ slot futures; best slot: per variate, the closest of the $K$ slot futures in hindsight. Left: the full bank vs.\ the same bank ranked by raw-window similarity. Right: ten per-variate slots vs.\ ten global slots (the ablation banks).}}
\usetikzlibrary{arrows.meta, positioning, shapes.multipart, calc, matrix}
\tikzset{
    methodbox/.style={
        draw,
        rounded corners,
        minimum width=3.8cm,
        minimum height=1.2cm,
        align=center,
        font=\small
    },
    methodarrow/.style={
        ->,
        thick
    }
}

\title{GNA: Granular Neighbor Assembly for Retrieval-Augmented Multivariate Time-Series Forecasting}

\author{Vincent Uhse \\
CERN \\
\texttt{vincent.uhse@cern.ch}}

\iclrfinalcopy

\newcommand{\ourmethod}{GNA}
\begin{document}

\maketitle
\lhead{Preprint. Under review.}

\begin{abstract}

\begin{cdraft}
Deep forecasters predict from a fixed-length lookback window, and lengthening it gives diminishing returns at a growing cost. Retrieval augmentation instead shows the model how similar past situations continued. Retrieving a whole past window gives every variate the continuation of the same past moment. In multivariate series, however, the best past match differs from variate to variate. We present \ourmethod{} (Granular Neighbor Assembly), a retrieval layer for forecasting backbones that assembles neighbors at two granularities: whole past windows, which keep the variates coherent, and per-variate neighbors, in which each variate takes its future from its own best-matching past. A learned gate decides, per forecast step and variate, how much to trust these futures against a persistence forecast, next to the backbone's own forecast. Candidates come from an embedding trained to predict each window's future, and retrieval is strictly causal: a past window is used only once its future has been observed. With the same lookback for every model and the same retrieval constants for all datasets, \ourmethod{} improves two Transformer backbones in 85 of 96 dataset--horizon settings, gives the lowest MSE on 8 of 12 standard benchmarks and beats its backbone in every seed on 10 of them. Both granularities are needed, and neighbors of mismatched queries are worse than none. Retrieval helps most where the lookback says least: the gate shifts trust to retrieved futures further ahead. Where it fails, on hourly non-stationary series at long horizons, the loss is consistent with a drifting level of the retrieved futures.
\end{cdraft}
\end{abstract}

\section{Introduction}
\label{sec:intro}
\begin{cdraft}
Multivariate time-series forecasting predicts future values of a set of variates from their recent history. It supports decisions wherever acting early pays off: weather forecasts are valued at \$31.5B per year for US households against \$5.1B in costs \citep{Lazo2009300BS}, and forecast errors are a standard signal for anomaly detection, for example in spacecraft telemetry \citep{hundman2018detecting}. Forecasting is hard because observations are noisy, variates interact, and the underlying system drifts, switches between operating regimes or enters situations that its recent past does not describe.

Deep forecasters \citep{nie2023a,liuITransformerInvertedTransformers2024,10.1609/aaai.v37i9.26317} predict from a lookback window of fixed length $L$. Anything that happened before the window is invisible to them. Lengthening the window has diminishing returns: for PatchTST, increasing the lookback from 336 to 720 steps changes the Traffic MSE from 0.367 to 0.365 and makes ETTh1 slightly worse at horizon 96 \citep[Table~9]{nie2023a}, while the cost of attention grows with $L$. No practical window reaches a situation last seen weeks earlier. In some applications a short lookback is also a requirement rather than a choice: after a restart or a change of operating conditions, only a short stretch of relevant past exists, and online monitoring limits latency and memory. Retrieval is a non-parametric alternative: find past windows that resemble the present one and show the model how they continued. This idea goes back to analog forecasting in meteorology \citep{lorenz1969atmospheric} and underlies retrieval-augmented language models \citep{NEURIPS2020_6b493230}.

Retrieval-augmented forecasters retrieve past windows whose histories resemble the query and pass their futures to the model; GTR \citep{cao2026enhancing} instead learns a global periodic memory. RAFT and RATD use a retrieved multivariate window as one unit \citep{han2025retrieval,NEURIPS2024_053ee34c}: all variates receive the continuation of the same past moment. In a network of hundreds of sensors, however, the past moment that best matches one sensor rarely matches all others, and the variates that fit the neighbor poorly receive a misleading future. Retrieving for each variate on its own, as foundation-model methods do \citep{ning2025tsrag,tireRetrievalAugmentedTime2026}, allows a closer match per variate but no longer keeps one past situation coherent across variates. Retrieved futures can also be useful for some forecast steps and misleading for others, so the model must learn when to use them.

We propose \ourmethod{} (Granular Neighbor Assembly, Fig.~\ref{fig:overview}), built on two ideas. First, neighbors are assembled at two granularities: \emph{global} slots keep one past situation coherent across variates, and \emph{per-variate} slots let each variate take its own best-matching past, so a slot can combine pieces of different past windows. Second, a learned gate weighs the slot futures against a persistence forecast per forecast step and variate, next to the backbone's own forecast, so the model can rely on the lookback for the next steps and on retrieved futures further out. Candidates come from a retrieval embedding that uses the futures of training windows as privileged information \citep{vapnik2009new}: it is trained to predict each window's future, while at test time it sees only the past. The bank is strictly causal and grows during deployment without retraining. \ourmethod{} needs only a backbone that produces a forecast and per-variate tokens; we use GridTST \citep{GridTST} and iTransformer \citep{liuITransformerInvertedTransformers2024}.

\begin{figure}[t]
\centering
\resizebox{\linewidth}{!}{\definecolor{dgInk}{HTML}{222222}
\definecolor{dgMuted}{HTML}{666666}
\definecolor{dgLight}{HTML}{E0E0E0}
\definecolor{dgBlue}{HTML}{1F77B4}
\definecolor{dgViolet}{HTML}{9467BD}
\definecolor{dgAqua}{HTML}{2CA02C}
\begin{tikzpicture}[
    font=\small,
    >={Stealth[length=2.2mm]},
    stage/.style={draw=dgMuted, rounded corners=2pt, fill=white, align=center,
                  minimum height=1.25cm, inner sep=3pt, line width=0.5pt},
    note/.style={font=\scriptsize, text=dgMuted, align=center},
    flow/.style={->, draw=dgInk, line width=0.6pt, rounded corners=4pt},
]
\begin{scope}[shift={(0,0)}]
  \draw[dgLight, fill=white, rounded corners=2pt] (-0.1,-0.75) rectangle (1.5,0.75);
  \draw[dgInk, line width=0.6pt] plot[smooth] coordinates {(0,0.45) (0.3,0.55) (0.6,0.4) (0.9,0.6) (1.2,0.5) (1.4,0.62)};
  \draw[dgInk, line width=0.6pt] plot[smooth] coordinates {(0,0.0) (0.3,-0.12) (0.6,0.08) (0.9,-0.05) (1.2,0.1) (1.4,0.0)};
  \draw[dgInk, line width=0.6pt] plot[smooth] coordinates {(0,-0.5) (0.3,-0.38) (0.6,-0.55) (0.9,-0.42) (1.2,-0.6) (1.4,-0.48)};
  \node[note, below] at (0.7,-0.75) {query window\\$\mathbf{X}_t$};
  \coordinate (q) at (1.5,0);
\end{scope}

\node[stage] (emb) at (3.15,0) {Future-aware\\embedder $\phi$};
\node[stage] (ret) at (6.2,0) {Causal retrieval\\{\scriptsize hybrid rank fusion}};
\node[stage] (asm) at (9.6,0) {Neighbor\\assembly};
\node[stage, draw=dgBlue, line width=0.8pt] (fc) at (12.35,0) {Retrieval-\\augmented\\forecaster};

\draw[flow] (q) -- (emb);
\draw[flow] (emb) -- node[note, above] {$\phi(\mathbf{X}_t)$} (ret);
\draw[flow] (ret) -- node[note, above] {candidates} (asm);
\draw[flow] (asm) -- node[note, above] {$K$ slots} (fc);
\draw[flow] (fc.east) -- ++(0.55,0) node[right, align=left] {$\hat{\mathbf{Y}}_t$};
\draw[flow] (0.7,0.75) -- (0.7,1.25) -| (fc.north);

\begin{scope}[shift={(3.35,-2.75)}]
  \draw[dgMuted, line width=0.5pt, ->] (0,0) -- (5.6,0) node[right, note] {time};
  \foreach \x/\y in {0.1/0.18, 1.0/0.42, 1.9/0.18, 2.8/0.42} {
    \fill[dgMuted!45] (\x,\y) rectangle ++(0.5,0.16);
    \fill[dgAqua!70] (\x+0.5,\y) rectangle ++(0.3,0.16);
  }
  \fill[dgInk!70] (4.2,0.3) rectangle ++(0.5,0.16);
  \draw[dgAqua!70, densely dashed] (4.7,0.3) rectangle ++(0.3,0.16);
  \draw[dgInk, line width=0.6pt] (4.7,-0.1) -- (4.7,0.75) node[above, note] {$t$};
  \node[note, anchor=north, align=center] at (1.9,-0.05) {bank of past windows (history {\color{dgMuted!45}\rule{6pt}{4pt}}, future {\color{dgAqua!70}\rule{6pt}{4pt}});\\eligible only if the future is fully observed by $t$};
  \node[note, anchor=west] at (5.05,0.38) {query};
  \coordinate (bank) at (2.85,0.7);
\end{scope}
\draw[flow] (bank) -- (ret.south);

\node[note, below=0.12cm of emb, align=center] {trained to predict\\the future; sees history\\only at test time};
\node[note, below=0.12cm of asm, align=left] {\textcolor{dgMuted}{\rule{5pt}{5pt}}\ global slots\\\textcolor{dgViolet}{\rule{5pt}{5pt}}\ per-variate slots};
\node[note, below=0.12cm of fc, text width=2.5cm] {learned gate + backbone\\+ cross-attention};
\end{tikzpicture}}
\caption{%
\cd{\ourmethod{} at forecast origin $t$. The future-aware embedder $\phi$ (Sec.~\ref{sec:embedder}) embeds the query window $\mathbf{X}_t$ (lines: variates). Causal retrieval (Sec.~\ref{sec:retrieval}) ranks the bank of past windows (bottom; grey: history, green: future); a window is eligible only if its future is fully observed by $t$ (Eq.~\ref{eq:eligibility}), and the query's own future (dashed) is unknown. Neighbor assembly (Sec.~\ref{sec:assembly}) fills $K$ slots with whole past windows (global, grey) or with a separate choice per variate (purple). The forecaster (Sec.~\ref{sec:forecaster}) combines the query window (top arrow) and the slots through a learned gate, the backbone and cross-attention into $\hat{\mathbf{Y}}_t$.}}
\label{fig:overview}
\end{figure}
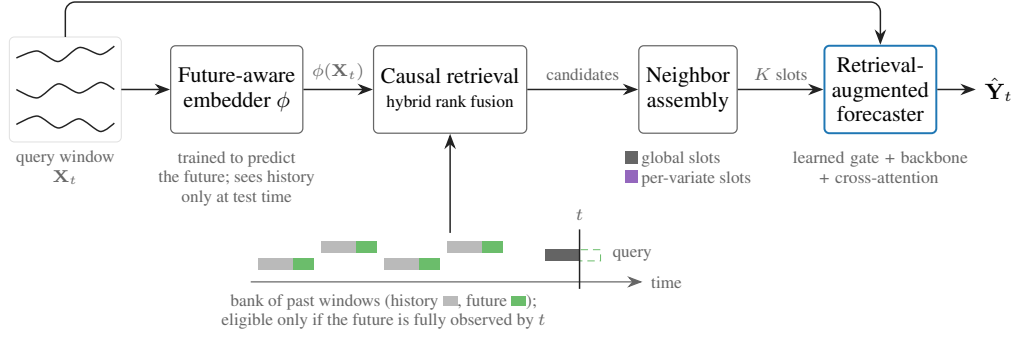

Our contributions are:
\begin{itemize}[leftmargin=*,itemsep=1pt,topsep=2pt]
\item \textbf{Granular neighbors with learned trust.} A retrieval layer that assembles neighbors at two granularities, whole past windows and single variates, and weighs them per forecast step and variate against persistence and the backbone (Sec.~\ref{sec:method}). To our knowledge, no previous retrieval-augmented forecaster combines whole-window and per-variate neighbors in one multivariate model (App.~\ref{app:related-comparison}).
\item \textbf{Evidence under a strict protocol.} Every model, including all baselines, uses the same lookback $L=96$; retrieval is strictly causal; the retrieval constants are shared by all datasets. \ourmethod{} improves GridTST in 43 and iTransformer in 42 of 48 dataset--horizon settings, gives the lowest MSE on 8 of 12 benchmarks, and beats its backbone in all three seeds on 10 of 12. Both granularities are needed, and replacing the retrieved windows by those of mismatched queries is worse than no retrieval, so the gain comes from relevant neighbors, not from added capacity (Secs.~\ref{sec:results}--\ref{sec:ablations}).
\item \textbf{When retrieval helps.} Retrieval helps most where the lookback says least: the gate moves trust from persistence to retrieved futures along the horizon. Where \ourmethod{} loses (hourly ETT data), the loss is a far-horizon effect consistent with level drift of the retrieved futures (Sec.~\ref{sec:when-helps}).
\end{itemize}
\end{cdraft}

\section{Related work}

\begin{cdraft}
\paragraph{Deep forecasters and the context window.}
Transformer forecasters differ in how they tokenize the lookback window: by time step \citep{zhouInformerEfficientTransformer2021,NEURIPS2021_bcc0d400}, by patch \citep{nie2023a}, by variate \citep{liuITransformerInvertedTransformers2024}, across both dimensions \citep{zhang2023crossformer,GridTST}, or with exogenous variables \citep{NEURIPS2024_0113ef46}. Linear and MLP models \citep{10.1609/aaai.v37i9.26317,wang2024timemixer,han2024softs,lin2024cyclenet,lin2025temporal}, convolutional \citep{wu2023timesnet,lstnet} and state-space models \citep{WANG2025129178} are strong alternatives, and instance normalization \citep{kim2022reversible} is standard. All of them condition only on the lookback window; \ourmethod{} adds information from outside it and uses two of them as backbones.

\paragraph{Retrieval-augmented forecasting.}
Nearest-neighbor and analog methods predict with the continuation of similar past states \citep{lorenz1969atmospheric,yakowitz1987nearest}. Recent deep variants retrieve (history, future) windows by the Pearson correlation of raw histories and aggregate the top matches (RAFT; \citealp{han2025retrieval}), use pre-trained encoder embeddings to condition a diffusion model (RATD; \citealp{NEURIPS2024_053ee34c}), or augment time-series foundation models with retrieved univariate segments (TS-RAG, RAF, TimeRAF; \citealp{ning2025tsrag,tireRetrievalAugmentedTime2026,zhang2024timeraf}). GTR \citep{cao2026enhancing} retrieves from a learned global-cycle embedding rather than from past windows. The multivariate methods treat each retrieved window as one unit, and the foundation-model methods retrieve for each univariate series on its own. \ourmethod{} combines both granularities in one multivariate model, trains its retrieval embedding to predict the future, and keeps the bank strictly causal while it grows at test time (App.~\ref{app:related-comparison} compares the mechanisms).

\paragraph{Learning the retriever.}
Retrieval-augmented language models \citep{NEURIPS2020_6b493230} can train the retriever from a downstream signal, for example by distilling the reader's attention into the retriever \citep{izacard:hal-03463398}. TimeRAF \citep{zhang2024timeraf} carries this over to forecasting and trains its retriever to rank higher the candidates that lower the forecaster's error. Our embedding uses the forecasting target more directly: it is trained to predict the future of each window, independently of the forecaster. Since futures are available only during training, this is learning with privileged information \citep{vapnik2009new}.
\end{cdraft}

\section{Method}
\label{sec:method}
\begin{cdraft}
\ourmethod{} turns retrieved past windows into \emph{slots}: candidate futures, one per variate, which a learned gate combines with a persistence forecast and the backbone's own forecast. Its core is how the slots are built, at two granularities (Sec.~\ref{sec:assembly}), and how much each is trusted, per forecast step and variate (Sec.~\ref{sec:forecaster}). A retrieval stage supplies the candidates (Sec.~\ref{sec:retrieval}). Figure~\ref{fig:overview} shows the stages for one query. The backbone sees exactly the lookback window it would see without retrieval, so \ourmethod{} works with any backbone that produces a forecast and per-variate tokens.
\end{cdraft}

\begin{figure}[t]
\centering
\resizebox{\linewidth}{!}{\definecolor{dgInk}{HTML}{222222}
\definecolor{dgMuted}{HTML}{666666}
\definecolor{dgLight}{HTML}{E0E0E0}
\definecolor{dgBlue}{HTML}{1F77B4}
\definecolor{dgViolet}{HTML}{9467BD}
\definecolor{dgAqua}{HTML}{2CA02C}
\begin{tikzpicture}[
    font=\small,
    >={Stealth[length=2mm]},
    box/.style={draw=dgMuted, rounded corners=2pt, fill=white, align=center,
                inner sep=3pt, line width=0.5pt, minimum height=0.7cm},
    op/.style={draw=dgMuted, circle, inner sep=1pt, fill=white, font=\small},
    note/.style={font=\scriptsize, text=dgMuted, align=center},
    flow/.style={->, draw=dgInk, line width=0.55pt, rounded corners=4pt},
    title/.style={font=\small, anchor=west, text=dgInk},
]
\def\cw{0.5}\def\ch{0.42}\def\top{0.8}
\filldraw[fill=dgLight!45, draw=dgMuted, rounded corners=6pt, line width=0.5pt] (-0.25,-0.62) rectangle (4.55,0.92);
\foreach \lab [count=\n from 0] in {a,b,c,d,e,f} {
  \pgfmathsetmacro{\x}{0.05+\n*0.66}
  \draw[dgMuted, rounded corners=1.5pt, line width=0.4pt, fill=white] (\x,-0.36) rectangle ({\x+0.52},0.36);
  \foreach \r in {0,1,2} {
    \fill[dgMuted!45] ({\x+0.05},{0.2-\r*0.2}) rectangle ({\x+0.32},{0.2-\r*0.2+0.1});
    \fill[dgAqua!70] ({\x+0.32},{0.2-\r*0.2}) rectangle ({\x+0.47},{0.2-\r*0.2+0.1});
  }
  \node[font=\scriptsize, text=dgInk] at ({\x+0.26},0.6) {\lab};
}
\node[font=\scriptsize, text=dgMuted] at (4.25,0) {$\cdots$};
\draw[dgMuted, ->, line width=0.4pt] (0.05,-0.5) -- (3.55,-0.5) node[right, note] {rank};
\node[note, anchor=north] at (2.15,-0.75) {candidate pool: hybrid rank fusion\\($\phi$, regime, seasonal, calendar);\\causally eligible, episode-diverse shortlist};
\def\gx{5.6}\def\px{8.35}
\foreach \r/\lab in {0/1, 1/2, 2/3} {\node[note, anchor=east] at ({\gx-0.12}, {\top-\r*\ch-0.5*\ch}) {var.\ \lab};}
\node[note, anchor=east] at ({\gx-0.12}, {\top-3*\ch-0.5*\ch}) {$\vdots$};
\node[note] at ({\gx+2.5*\cw}, {\top+0.28}) {global slots};
\node[note] at ({\px+2.5*\cw}, {\top+0.28}) {per-variate slots};
\foreach \c/\src in {0/a, 1/b, 2/c, 3/d, 4/e} {
  \foreach \r in {0,1,2,3} {
    \filldraw[fill=dgMuted!25, draw=white, line width=0.8pt]
      ({\gx+\c*\cw}, {\top-\r*\ch}) rectangle ({\gx+(\c+1)*\cw}, {\top-(\r+1)*\ch});
    \node[font=\scriptsize, text=dgInk] at ({\gx+\c*\cw+0.5*\cw}, {\top-\r*\ch-0.5*\ch}) {\src};
  }
}
\foreach \srcs [count=\r from 0] in {{c,f,a,g,d}, {b,g,e,a,f}, {f,c,g,b,e}, {a,e,d,g,c}} {
  \foreach \s [count=\c from 0] in \srcs {
    \filldraw[fill=dgViolet!22, draw=white, line width=0.8pt]
      ({\px+\c*\cw}, {\top-\r*\ch}) rectangle ({\px+(\c+1)*\cw}, {\top-(\r+1)*\ch});
    \node[font=\scriptsize, text=dgInk] at ({\px+\c*\cw+0.5*\cw}, {\top-\r*\ch-0.5*\ch}) {\s};
  }
}
\draw[flow] (3.9,0.92) -- (3.9,1.45) -- ({\gx+2.5*\cw},1.45) -- ({\gx+2.5*\cw},{\top+0.45});
\node[note, anchor=south west, align=left] at (4.05,1.47) {top-ranked windows\\taken for all variates};
\draw[flow] (0.9,0.92) -- (0.9,2.25) -- ({\px+2.5*\cw},2.25) -- ({\px+2.5*\cw},{\top+0.45});
\node[note, anchor=south west, align=left] at (1.05,2.27) {per variate: re-ranked by past-shape fit to the query\\(second pool, candidates $\geq$ 96 steps apart); top 5 taken};
\node[note, anchor=north] at ({\gx+5.5*\cw}, -1.05) {each cell: that variate's {\color{dgAqua}future} over the $H$ forecast steps\\(letters: source window)};

\end{tikzpicture}}
\caption{%
\cd{Neighbor assembly for one query. Left: fused candidates in rank order; letters name source windows, and each box shows one row per variate (grey: history, green: future). Middle: the global slots use the five top-ranked windows for all variates. Right: for the per-variate slots, each variate re-ranks a second, spaced candidate pool by how well its own past matches the query's past and takes its top five, so one column combines different source windows (variate 1: c, f, a, g, d). Each cell holds that variate's $H$-step future; the $K=10$ columns are the slot futures $\mathbf{c}_1,\dots,\mathbf{c}_K$ (Eq.~\ref{eq:slot}).}}
\label{fig:assembly}
\end{figure}
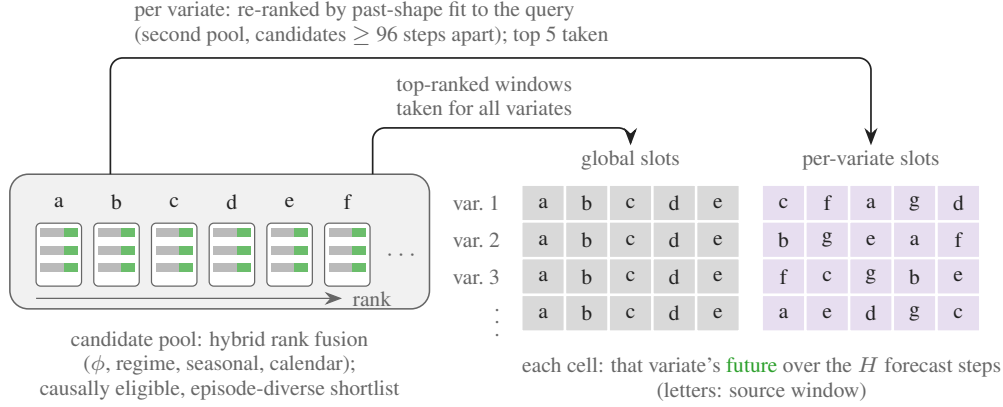

\subsection{Problem setup and causal eligibility}
\begin{cdraft}
We observe a multivariate series $\mathbf{x}_t \in \mathbb{R}^D$ with $D$ variates. At forecast origin $t$, the query window is $\mathbf{X}_t = \mathbf{x}_{t-L+1:t} \in \mathbb{R}^{L\times D}$ and the target is $\mathbf{Y}_t = \mathbf{x}_{t+1:t+H} \in \mathbb{R}^{H\times D}$, with lookback $L$ and horizon $H$; $h \in \{1,\dots,H\}$ denotes the forecast step. The bank contains every window $(\mathbf{X}_s,\mathbf{Y}_s)$ of the series, indexed by its origin $s$. A bank window may serve query $t$ only if
\begin{equation}
s \le t - \max(L, H).
\label{eq:eligibility}
\end{equation}
Then its future ends at $s+H \le t$, so it has been fully observed at the forecast origin, and its past ends at $s \le t-L$, so it does not overlap the query's past. The rule holds for training, validation and test queries alike: a training query retrieves only earlier training windows, and at test time the bank grows as new futures are observed, without retraining (App.~\ref{app:method-details}, Fig.~\ref{fig:pipeline}). We audited all banks for this rule. Table~\ref{tab:notation} (App.~\ref{app:method-details}) lists all symbols.
\end{cdraft}

\subsection{Retrieving candidates}
\label{sec:embedder}\label{sec:retrieval}
\begin{cdraft}
Retrieval should return windows whose futures resemble the query's unknown future, so we train the retrieval embedding with the futures of training windows as targets. A teacher $\phi_T$, an MLP on the flattened window, maps window $i$ to $\mathbf{z}_i = \phi_T(\mathbf{X}_i) \in \mathbb{R}^{d_z}$, and a head $\psi$, used only in training, predicts the window's future from it:
\begin{equation}
\mathcal{L}_T = \mathrm{MSE}\big(\psi(\mathbf{z}_i), \mathbf{Y}_i\big).
\label{eq:teacher}
\end{equation}
The embedding thus keeps what the past window says about its future. Futures appear only in the loss, so the embedder needs only the past at test time; this is learning with privileged information \citep{vapnik2009new}. Retrieval uses a small Transformer $\phi$ distilled from the teacher (App.~\ref{app:method-details}, Fig.~\ref{fig:embedder}). For query $t$ we search all eligible windows (Eq.~\ref{eq:eligibility}) exactly, by cosine similarity of their $\phi$ embeddings. A learned embedding can miss coarse context such as the operating level or the time of day, so we combine its ranking with three simple context signals $j$: the similarity of low-dimensional regime features, lags at which the training series is periodic, and calendar match. With $\bar r(s)\in[0,1]$ the normalized rank of candidate $s$ under each signal, candidates are ordered by
\begin{equation}
\rho(s) = \bar r_\phi(s) + \textstyle\sum_j w_j\,\bar r_j(s),
\label{eq:fusion}
\end{equation}
lowest first, with fixed weights shared by all datasets (App.~\ref{app:method-details}); PEMS and Solar have no time stamps, so the calendar signal is not used there. The search keeps $C=200$ candidates, spread over different episodes so that one episode cannot fill the list.
\end{cdraft}

\subsection{Granular neighbor assembly}
\label{sec:assembly}
\begin{cdraft}
Multivariate retrieval forecasters such as RAFT and RATD treat a retrieved window as one unit: every variate receives the future of the same past window. In a network of hundreds of sensors, however, the past moment that best matches one sensor rarely best matches all others. \ourmethod{} therefore fills $K$ \emph{slots}, in which each variate can have its own source. Slot $k$ assigns to every variate $d$ an eligible source window $s_{k,d}$ and provides that variate's past and future, scaled with the mean $\mu_{t,d}$ and standard deviation $\sigma_{t,d}$ of variate $d$ in the query window:
\begin{equation}
\mathbf{c}_k[:,d] = \big(\mathbf{Y}_{s_{k,d}}[:,d] - \mu_{t,d}\big) / \sigma_{t,d} \;\in \mathbb{R}^{H}.
\label{eq:slot}
\end{equation}
We use two kinds of slots (Fig.~\ref{fig:assembly}). The five \emph{global} slots take the five top-ranked candidates for all variates ($s_{k,d}=s_k$). They keep one past situation coherent across variates, such as a network-wide rush hour. The five \emph{per-variate} slots choose $s_{k,d}$ separately for each variate. A second fused list, built as above but with candidates at least $L$ steps apart, provides a pool of $M=50$ candidates. Variate $d$ ranks the candidates by how well \emph{its own} past matches the query's past, and takes its top five. The match is a fixed multi-scale distance between the two univariate pasts (App.~\ref{app:method-details}). A per-variate slot is thus a composite that never occurred as one multivariate window. Global slots preserve cross-variate structure, while per-variate slots find a closer match when variates decouple. The spacing keeps per-variate slots from being near-copies of one episode.
\end{cdraft}

\begin{figure}[t]
\centering
\resizebox{\linewidth}{!}{\definecolor{dgInk}{HTML}{222222}
\definecolor{dgMuted}{HTML}{666666}
\definecolor{dgLight}{HTML}{E0E0E0}
\definecolor{dgBlue}{HTML}{1F77B4}
\definecolor{dgViolet}{HTML}{9467BD}
\definecolor{dgAqua}{HTML}{2CA02C}
\begin{tikzpicture}[
    font=\small,
    >={Stealth[length=2mm]},
    box/.style={draw=dgMuted, rounded corners=2pt, fill=white, align=center,
                inner sep=3pt, line width=0.5pt, minimum height=0.7cm},
    op/.style={draw=dgMuted, circle, inner sep=1pt, fill=white, font=\small},
    note/.style={font=\scriptsize, text=dgMuted, align=center},
    flow/.style={->, draw=dgInk, line width=0.55pt, rounded corners=4pt},
    title/.style={font=\small, anchor=west, text=dgInk},
]
\node[box, text width=1.7cm] (x) at (-0.4,0.55) {query window $\mathbf{X}_t$};
\node[box, text width=2.3cm, draw=dgViolet] (nb) at (-0.4,-1.8) {$K$ slots:\\{\color{dgAqua}futures}\\{\scriptsize (+ pasts for gate fit)}};
\draw[flow, draw=dgViolet] (x.south) -- node[note, left, align=right] {retrieval\\+ assembly} (nb.north);
\node[box, text width=3.9cm] (gate) at (3.7,1.2) {gate $g_k$ per step \& variate\\{\scriptsize from past fit, slot type, calendar}};
\node[box, text width=2.4cm] (mix) at (8.35,1.2) {$\sum_{k=0}^{K} g_k\,{\color{dgAqua}\mathbf{c}_k}$\\{\scriptsize $\mathbf{c}_0$: persistence, $\mathbf{c}_{1..K}$: {\color{dgAqua}futures}}};
\node[box, draw=dgBlue, line width=0.8pt, text width=3.9cm] (bb) at (3.7,-0.1) {backbone $f$ (GridTST or iTransformer)};
\node[box, text width=2.4cm] (bbo) at (8.35,-0.1) {$\alpha\, f(\mathbf{X}_t)$};
\node[box, text width=3.9cm, draw=dgAqua] (ftok) at (3.7,-1.8) {neighbor-{\color{dgAqua}future} tokens\\{\scriptsize with slot-type embedding}};
\node[box, text width=2.4cm] (ca) at (8.35,-1.8) {cross-attention\\{\scriptsize one query per variate}};
\node[box] (proj) at (10.75,-1.8) {$\beta\, R(\cdot)$};
\node[op] (sum) at (12.0,-0.1) {$+$};
\node[box, text width=1.3cm] (out) at (13.4,-0.1) {denorm.\\$\hat{\mathbf{Y}}_t$};
\draw[flow] (x.east) -- ++(0.3,0) |- (gate.west);
\draw[flow] (x.east) -- ++(0.3,0) |- (bb.west);
\draw[flow] (nb.east) -- ++(0.45,0) |- ([yshift=-5pt]gate.west);
\draw[flow] (nb.east) -- (ftok.west);
\draw[flow] (gate) -- (mix);
\draw[flow] (bb) -- (bbo);
\draw[flow] (ftok) -- node[note, below, align=center] {keys,\\values} (ca);
\draw[flow] ([xshift=1.2cm]bb.south) -- ++(0,-0.28) -| node[note, pos=0.25, above] {variate tokens} ([xshift=-0.6cm]ca.north);
\draw[flow] (ca) -- (proj);
\draw[flow] (mix.east) -| (sum.north);
\draw[flow] (bbo) -- (sum);
\draw[flow] (proj.east) -| (sum.south);
\draw[flow] (sum) -- (out);
\node[note, anchor=west] at (9.65,0.25) {$\alpha, \beta$ learned};

\end{tikzpicture}}
\caption{%
\cd{The retrieval-augmented forecaster (Eq.~\ref{eq:forecast}). Top: the gate compares each slot's past with the query's past per variate and, with slot type and calendar match, weights persistence $\mathbf{c}_0$ and the slot futures $\mathbf{c}_{1..K}$ per forecast step and variate. Middle: the backbone $f$ forecasts from $\mathbf{X}_t$ alone, scaled by $\alpha$. Bottom: each variate token of the backbone attends to its $K$ slot-future tokens, and the result is projected to $H$ steps and scaled by $\beta$. The sum is denormalized to $\hat{\mathbf{Y}}_t$.}}
\label{fig:forecaster}
\end{figure}
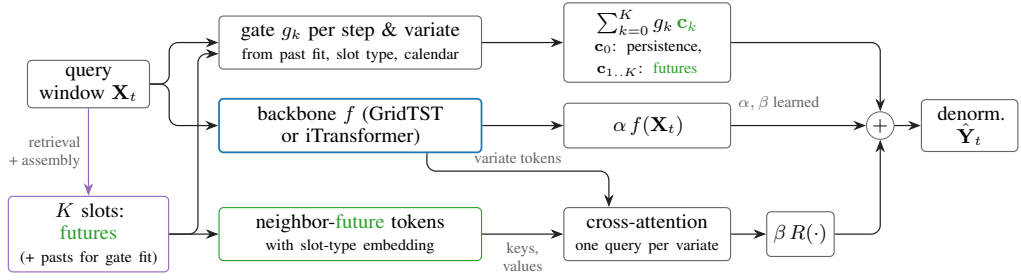
\subsection{Forecasting with learned trust}
\label{sec:forecaster}
\begin{cdraft}
Let $\mathbf{c}_1,\dots,\mathbf{c}_K \in \mathbb{R}^{H\times D}$ be the slot futures of Eq.~\ref{eq:slot} and $\mathbf{c}_0$ the persistence forecast, i.e.\ the query's last value repeated $H$ times, all in the query's normalized space. The forecast is
\begin{equation}
\hat{\mathbf{Y}}_t = \operatorname{denorm}\Big(\sum_{k=0}^{K} \mathbf{g}_k \odot \mathbf{c}_k \;+\; \alpha\, f(\mathbf{X}_t) \;+\; \beta\, R\big(\mathbf{X}_t, \mathbf{c}_{1:K}\big)\Big),
\label{eq:forecast}
\end{equation}
where $f$ is the backbone, $\alpha$ and $\beta$ are learned scalars initialized at 0.1, and $\operatorname{denorm}$ inverts the query's instance normalization \citep{kim2022reversible} (Fig.~\ref{fig:forecaster}). The first term is a learned trust over the candidates. The gate weights $\mathbf{g}_k \in \mathbb{R}^{H\times D}$ sum to one over $k=0,\dots,K$ at every forecast step and variate. They are computed by a small MLP from how well slot $k$'s past fits the query's past, for that variate and averaged over all variates, and from the slot type and calendar match (App.~\ref{app:method-details}). Persistence has its own learned logit per forecast step, so where no slot fits, the gate can fall back to it. The weights differ per forecast step, so the gate can trust persistence for the next few steps and retrieved futures further out. The second term keeps the backbone's own forecast. The third term, $R$, lets the backbone condition on the neighbors: each variate token of $f$ attends to the $K$ slot-future tokens of that variate (with a slot-type embedding), and the result is projected to $H$ steps.
\end{cdraft}

\subsection{Training and cost}
\label{sec:training}
\begin{cdraft}
Training runs in four stages (Fig.~\ref{fig:pipeline}): teacher embedder, distilled student, bank construction (embed all windows, rank the eligible candidates of every query, assemble the slots), and forecaster. The forecaster is trained end to end with MSE, with the embedder and bank fixed, using the backbone's training recipe for 10 epochs with batch size 32 (8 for Traffic): AdamW with weight decay 0.01 and a one-cycle schedule for GridTST, and Adam with the learning rate halved after every epoch for iTransformer. Exact search costs $O(N(d_z+d_r))$ per query for a bank of $N$ windows, so searching for all queries takes $O(N^2(d_z+d_r))$ time, the per-variate re-ranking adds $O(NMLD)$, and the searched embeddings need $O(N(d_z+d_r))$ memory. On the four datasets timed on an otherwise idle GPU (7 to 321 variates), the bank is built in 3--7 minutes, the index takes 0.15--0.40\,GB, training is 14--51\% slower per epoch than the backbone alone, and online retrieval takes 19--25\,ms per query on a CPU (App.~\ref{app:compute}).
\end{cdraft}

\section{Experimental setup}
\label{sec:setup}
\begin{cdraft}
Our experiments answer three questions: does retrieval improve forecasts under a strict same-lookback protocol (Sec.~\ref{sec:results}), which components carry the gain (Sec.~\ref{sec:ablations}), and when does retrieval help or fail (Sec.~\ref{sec:when-helps})?

\paragraph{Datasets.} We use twelve standard benchmarks (App.~\ref{app:datasets}): the traffic sensor networks PEMS03/04/07/08 (as processed by \citealp{NEURIPS2022_266983d0}) with horizons $H\in\{12,24,48,96\}$; ETTh1/h2 and ETTm1/m2 \citep{zhouInformerEfficientTransformer2021}; and ECL, Traffic, Weather \citep{NEURIPS2021_bcc0d400} and Solar \citep{lstnet}, all with $H\in\{96,192,336,720\}$ and the standard chronological splits.

\paragraph{Baselines and protocol.} Every model uses lookback $L=96$. Baselines are the two backbones without retrieval, the retrieval-based GTR \citep{cao2026enhancing} and RAFT \citep{han2025retrieval}, and 14 further recent forecasters in App.~\ref{app:full-results}. We take published numbers at $L=96$; where a method's own paper uses another lookback, we use $L=96$ numbers reported by a third party (marked $\ddagger$). GridTST has no published $L=96$ results, so we run it ourselves, with the same code, training recipe and checkpoint selection as \ourmethod{}: without retrieval, Eq.~\ref{eq:forecast} reduces to $\operatorname{denorm}(\mathbf{c}_0 + \alpha f(\mathbf{X}_t))$, i.e.\ persistence plus the scaled backbone forecast. We ran no hyperparameter search: GridTST architectures follow the official configurations, the retrieval constants (fusion and re-rank weights, $K$, spacing) are untuned defaults shared by all datasets, and the few per-dataset settings listed in App.~\ref{app:hyperparameters} were set by hand. Tables report seed 2023; App.~\ref{app:seeds} repeats all datasets with three seeds.

\paragraph{Metrics.} We report MSE and MAE on standardized data, averaged over the four horizons (per horizon in App.~\ref{app:full-results}), on the full test set (no dropped last batch). Every run keeps the checkpoint with the lowest validation MSE over its 10 epochs.
\end{cdraft}

\section{Results}
\label{sec:results}

\providecommand{\ourmethod}{GNA}
\begin{table*}[t]
\centering
\scriptsize
\setlength{\tabcolsep}{3pt}
\begin{tabular}{lcccccccccccc}
\toprule
 & \multicolumn{4}{c}{\textit{\ourmethod{} (this work)}} & \multicolumn{4}{c}{\textit{Backbones}} & \multicolumn{4}{c}{\textit{Retrieval-based}} \\
\cmidrule(lr){2-5}\cmidrule(lr){6-9}\cmidrule(lr){10-13}
Dataset & \multicolumn{2}{c}{w/ GridTST} & \multicolumn{2}{c}{w/ iTransformer} & \multicolumn{2}{c}{GridTST} & \multicolumn{2}{c}{iTransformer} & \multicolumn{2}{c}{GTR} & \multicolumn{2}{c}{RAFT$^{\ddagger}$} \\
 & \multicolumn{2}{c}{\tiny } & \multicolumn{2}{c}{\tiny } & \multicolumn{2}{c}{\tiny arXiv \citeyear{GridTST}, our run} & \multicolumn{2}{c}{\tiny ICLR \citeyear{liuITransformerInvertedTransformers2024}} & \multicolumn{2}{c}{\tiny ICLR \citeyear{cao2026enhancing}} & \multicolumn{2}{c}{\tiny ICML \citeyear{han2025retrieval}} \\
\cmidrule(lr){2-3}\cmidrule(lr){4-5}\cmidrule(lr){6-7}\cmidrule(lr){8-9}\cmidrule(lr){10-11}\cmidrule(lr){12-13}
 & MSE & MAE & MSE & MAE & MSE & MAE & MSE & MAE & MSE & MAE & MSE & MAE \\
\midrule
PEMS03 & \textbf{\textcolor{red}{0.082}} & \textbf{\textcolor{red}{0.182}} & \underline{\textcolor{blue}{0.084}} & \underline{\textcolor{blue}{0.183}} & 0.153 & 0.260 & 0.113 & 0.221 & 0.087 & 0.189 & 0.144 & 0.230 \\
PEMS04 & \textbf{\textcolor{red}{0.080}} & \textbf{\textcolor{red}{0.182}} & \underline{\textcolor{blue}{0.083}} & \underline{\textcolor{blue}{0.184}} & 0.174 & 0.283 & 0.111 & 0.221 & 0.087 & 0.189 & 0.104 & 0.210 \\
PEMS07 & \textbf{\textcolor{red}{0.065}} & \textbf{\textcolor{red}{0.155}} & \underline{\textcolor{blue}{0.067}} & \underline{\textcolor{blue}{0.156}} & 0.134 & 0.241 & 0.101 & 0.204 & 0.076 & 0.169 & 0.094 & 0.193 \\
PEMS08 & \textbf{\textcolor{red}{0.076}} & \textbf{\textcolor{red}{0.176}} & \textbf{\textcolor{red}{0.076}} & \textbf{\textcolor{red}{0.176}} & 0.184 & 0.275 & 0.150 & 0.226 & 0.142 & 0.222 & 0.151 & 0.234 \\
\midrule
ETTh1 & 0.463 & 0.459 & 0.473 & 0.464 & 0.456 & 0.448 & 0.454 & 0.447 & 0.439 & 0.434 & \textbf{\textcolor{red}{0.428}} & \underline{\textcolor{blue}{0.433}} \\
ETTh2 & 0.420 & 0.437 & 0.427 & 0.441 & 0.395 & 0.412 & 0.383 & 0.407 & 0.372 & 0.400 & 0.382 & 0.410 \\
ETTm1 & \textbf{\textcolor{red}{0.359}} & \textbf{\textcolor{red}{0.386}} & \underline{\textcolor{blue}{0.365}} & \underline{\textcolor{blue}{0.389}} & 0.386 & 0.402 & 0.407 & 0.410 & 0.367 & \underline{\textcolor{blue}{0.389}} & 0.381 & 0.400 \\
ETTm2 & 0.272 & 0.329 & 0.278 & 0.331 & 0.287 & 0.332 & 0.288 & 0.332 & \underline{\textcolor{blue}{0.268}} & \underline{\textcolor{blue}{0.315}} & 0.281 & 0.330 \\
ECL & \textbf{\textcolor{red}{0.153}} & \textbf{\textcolor{red}{0.252}} & \underline{\textcolor{blue}{0.154}} & \underline{\textcolor{blue}{0.255}} & 0.174 & 0.266 & 0.178 & 0.270 & 0.166 & 0.260 & 0.175 & 0.272 \\
Solar & \underline{\textcolor{blue}{0.198}} & \underline{\textcolor{blue}{0.245}} & 0.201 & \textbf{\textcolor{red}{0.243}} & 0.228 & 0.278 & 0.233 & 0.262 & \textbf{\textcolor{red}{0.194}} & \underline{\textcolor{blue}{0.245}} & 0.301 & 0.303 \\
Traffic & \textbf{\textcolor{red}{0.397}} & 0.269 & \underline{\textcolor{blue}{0.409}} & \textbf{\textcolor{red}{0.263}} & 0.412 & 0.271 & 0.428 & 0.282 & 0.470 & 0.280 & 0.414 & 0.284 \\
Weather & \textbf{\textcolor{red}{0.235}} & 0.270 & \underline{\textcolor{blue}{0.239}} & 0.273 & 0.245 & 0.272 & 0.258 & 0.278 & \underline{\textcolor{blue}{0.239}} & \textbf{\textcolor{red}{0.268}} & 0.270 & 0.309 \\
\midrule
1st count & \textbf{\textcolor{red}{8}} & \textbf{\textcolor{red}{6}} & 1 & 3 & 0 & 0 & 0 & 0 & 1 & 1 & 1 & 0 \\
\bottomrule
\end{tabular}
\par\medskip
\begin{tabular}{lcccccccccccccc}
\toprule
 & \multicolumn{14}{c}{\textit{Other forecasters}} \\
\cmidrule(lr){2-15}
Dataset & \multicolumn{2}{c}{TQNet} & \multicolumn{2}{c}{Amplifier} & \multicolumn{2}{c}{CycleNet} & \multicolumn{2}{c}{SOFTS} & \multicolumn{2}{c}{TimeXer} & \multicolumn{2}{c}{TimeMixer} & \multicolumn{2}{c}{PatchTST$^{\ddagger}$} \\
 & \multicolumn{2}{c}{\tiny ICML \citeyear{lin2025temporal}} & \multicolumn{2}{c}{\tiny AAAI \citeyear{10.1609/aaai.v39i11.33267}} & \multicolumn{2}{c}{\tiny NeurIPS \citeyear{lin2024cyclenet}} & \multicolumn{2}{c}{\tiny NeurIPS \citeyear{han2024softs}} & \multicolumn{2}{c}{\tiny NeurIPS \citeyear{NEURIPS2024_0113ef46}} & \multicolumn{2}{c}{\tiny ICLR \citeyear{wang2024timemixer}} & \multicolumn{2}{c}{\tiny ICLR \citeyear{nie2023a}} \\
\cmidrule(lr){2-3}\cmidrule(lr){4-5}\cmidrule(lr){6-7}\cmidrule(lr){8-9}\cmidrule(lr){10-11}\cmidrule(lr){12-13}\cmidrule(lr){14-15}
 & MSE & MAE & MSE & MAE & MSE & MAE & MSE & MAE & MSE & MAE & MSE & MAE & MSE & MAE \\
\midrule
PEMS03 & 0.097 & 0.203 & 0.131$^{\ddagger}$ & 0.239$^{\ddagger}$ & 0.118$^{\ddagger}$ & 0.226$^{\ddagger}$ & 0.104 & 0.210 & 0.112$^{\ddagger}$ & 0.214$^{\ddagger}$ & 0.167$^{\ddagger}$ & 0.267$^{\ddagger}$ & 0.180 & 0.291 \\
PEMS04 & 0.091 & 0.197 & 0.135$^{\ddagger}$ & 0.249$^{\ddagger}$ & 0.119$^{\ddagger}$ & 0.232$^{\ddagger}$ & 0.102 & 0.208 & 0.105$^{\ddagger}$ & 0.209$^{\ddagger}$ & 0.185$^{\ddagger}$ & 0.287$^{\ddagger}$ & 0.195 & 0.307 \\
PEMS07 & 0.075 & 0.171 & 0.122$^{\ddagger}$ & 0.226$^{\ddagger}$ & 0.113$^{\ddagger}$ & 0.214$^{\ddagger}$ & 0.087 & 0.184 & 0.085$^{\ddagger}$ & 0.182$^{\ddagger}$ & 0.181$^{\ddagger}$ & 0.271$^{\ddagger}$ & 0.211 & 0.303 \\
PEMS08 & 0.142 & 0.229 & 0.183$^{\ddagger}$ & 0.271$^{\ddagger}$ & 0.150$^{\ddagger}$ & 0.246$^{\ddagger}$ & \underline{\textcolor{blue}{0.138}} & \underline{\textcolor{blue}{0.219}} & 0.175$^{\ddagger}$ & 0.250$^{\ddagger}$ & 0.226$^{\ddagger}$ & 0.299$^{\ddagger}$ & 0.280 & 0.321 \\
\midrule
ETTh1 & 0.441 & 0.434 & \underline{\textcolor{blue}{0.430}} & \textbf{\textcolor{red}{0.428}} & 0.457 & 0.441 & 0.449 & 0.442 & 0.437 & 0.437 & 0.447 & 0.440 & 0.469 & 0.454 \\
ETTh2 & 0.378 & 0.402 & \textbf{\textcolor{red}{0.359}} & \textbf{\textcolor{red}{0.391}} & 0.388 & 0.409 & 0.373 & 0.400 & 0.367 & 0.396 & \underline{\textcolor{blue}{0.364}} & \underline{\textcolor{blue}{0.395}} & 0.387 & 0.407 \\
ETTm1 & 0.377 & 0.393 & 0.381 & 0.394 & 0.379 & 0.396 & 0.393 & 0.403 & 0.382 & 0.397 & 0.381 & 0.395 & 0.387 & 0.400 \\
ETTm2 & 0.277 & 0.323 & 0.276 & 0.323 & \textbf{\textcolor{red}{0.266}} & \textbf{\textcolor{red}{0.314}} & 0.287 & 0.330 & 0.274 & 0.322 & 0.275 & 0.323 & 0.281 & 0.326 \\
ECL & 0.164 & 0.259 & 0.171 & 0.265 & 0.168 & 0.259 & 0.174 & 0.264 & 0.171 & 0.270 & 0.182 & 0.272 & 0.205 & 0.290 \\
Solar & \underline{\textcolor{blue}{0.198}} & 0.256 & 0.241$^{\ddagger}$ & 0.270$^{\ddagger}$ & 0.210 & 0.261 & 0.229 & 0.256 & 0.237$^{\ddagger}$ & 0.302$^{\ddagger}$ & 0.216 & 0.280 & 0.270 & 0.307 \\
Traffic & 0.445 & 0.276 & 0.482 & 0.315 & 0.472 & 0.301 & \underline{\textcolor{blue}{0.409}} & \underline{\textcolor{blue}{0.267}} & 0.466 & 0.287 & 0.484 & 0.297 & 0.481 & 0.304 \\
Weather & 0.242 & \underline{\textcolor{blue}{0.269}} & 0.243 & 0.271 & 0.243 & 0.271 & 0.255 & 0.278 & 0.241 & 0.271 & 0.240 & 0.271 & 0.259 & 0.281 \\
\midrule
1st count & 0 & 0 & 1 & 2 & 1 & 1 & 0 & 0 & 0 & 0 & 0 & 0 & 0 & 0 \\
\bottomrule
\end{tabular}
\caption{Results with lookback 96, averaged over horizons (PEMS: 12/24/48/96; others: 96/192/336/720). Baselines from the respective publications, except GridTST (our run; no published lookback-96 results). $^{\ddagger}$Lookback-96 result reported by a third party (sources in App.~\ref{app:full-results}). Best in red bold, second underlined.}
\label{tab:main_comparison_table}
\end{table*}

\begin{cdraft}
Table~\ref{tab:main_comparison_table} compares \ourmethod{} with its backbones and with retrieval-based forecasters. With GridTST, \ourmethod{} has the lowest MSE on 8 of the 12 datasets, also among all 18 baselines in App.~\ref{app:full-results}. The gains are largest on the PEMS sensor networks, where \ourmethod{} roughly halves the MSE of the plain GridTST and is 5--44\% below the best baseline (here and below: the best of all 18 baselines). With a lookback of 96 five-minute steps (8 hours), these models see a third of the daily cycle, and retrieved windows supply how comparable past situations continued. The gain is more than a recovered daily cycle: GTR, CycleNet and TQNet model periodic structure explicitly, and \ourmethod{} is 5--49\% below each of them on every PEMS dataset. Likewise, ETTm windows cover a full day and ECL windows four days, yet \ourmethod{} lowers the MSE of GridTST by 7\% on ETTm1 and 12\% on ECL. On ECL, Traffic, ETTm1 and Weather, \ourmethod{} is 0.5--4.9\% below the best baseline. Among retrieval-based forecasters, it is better than GTR on 8 of 12 datasets and better than RAFT on 10 of 12. It is not the best on ETTm2 and Solar (2.9\% and 2.1\% above the best), and on ETTh1 and ETTh2 it is worse than its own backbone; Sec.~\ref{sec:when-helps} analyzes why.

\paragraph{Retrieval vs.\ no retrieval.} The cleanest comparison holds the backbone architecture, data and lookback fixed and adds retrieval (App.~\ref{app:full-results}, Table~\ref{tab:retrieval-vs-baseline}). Retrieval improves GridTST in 43 and iTransformer in 42 of 48 dataset--horizon settings. Over three seeds, \ourmethod{} beats the plain GridTST in every seed on 10 of 12 datasets (not on ETTh1/h2), and its seed-to-seed variation is small (median 0.3\% of the MSE, vs.\ 4\% for the plain backbone; App.~\ref{app:seeds}). Outside PEMS and ETTh, the seed-mean gains over GridTST are 4--21\%, far above this variation. The gain is not explained by capacity: with iTransformer, \ourmethod{} uses a smaller iTransformer than the plain baseline and, including the retrieval embedder $\phi$, has fewer parameters in all 48 settings (App.~\ref{app:hyperparameters}), and with mismatched neighbors, the same architecture performs worse than without retrieval (Sec.~\ref{sec:ablations}).
\end{cdraft}

\section{Ablations and sensitivity}
\label{sec:ablations}
\begin{figure}[t]
\centering
\includegraphics{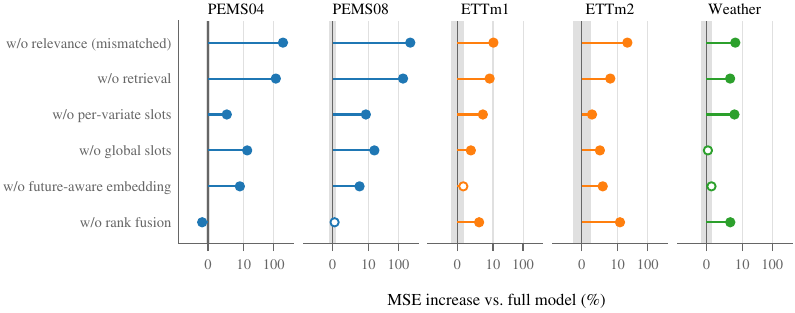}
\caption{%
\cd{Ablations (GridTST backbone): change in horizon-averaged test MSE relative to the full \ourmethod{}; positive = worse without the component. Filled: larger than twice the full model's seed standard deviation on that dataset (three seeds); hollow: within this noise. Symmetric-log $x$ axis (linear within $\pm5\%$); ablation runs use one seed.}}
\label{fig:ablation-dots}
\end{figure}

\begin{cdraft}
We ablate on five datasets (PEMS04, PEMS08, ETTm1, ETTm2, Weather) with the GridTST backbone (Fig.~\ref{fig:ablation-dots}; table in App.~\ref{app:ablations}). \emph{Is the gain due to retrieval?} Feeding each query the neighbors of a different query (mismatched), with the same architecture trained and tested this way, raises the MSE by 6--247\% (median 21\%), more than removing retrieval (4--142\%, median 8\%), on all five datasets. Retrieved futures therefore help only when they are relevant, and the model cannot fully ignore irrelevant ones (Sec.~\ref{sec:limitations}). \emph{Are both granularities needed?} Using only global slots raises the MSE by 1.9--8.3\% (mean 4.8\%), using only per-variate slots by 0.3--15.8\% (mean 7.0\%), so the mixture beats either slot type alone. Measured on the retrieved futures themselves, the best per-variate slot is 17--45\% closer to the true future than the best global slot on all five datasets (App.~\ref{app:ablations}). \emph{How much does the retrieval stage add?} Its parts add smaller, dataset-dependent gains. Ranking with standardized raw windows instead of the learned embedding raises the MSE by 0.9--7.4\% (mean 3.7\%). Ranking by the embedding alone, without the context signals, raises it where calendar and regime structure is informative (ETTm1, ETTm2, Weather: $+4.1$, $+12.0$, $+4.5\%$), but not on the PEMS sensors (PEMS08: $+0.4\%$, PEMS04: $-1.1\%$). \emph{How many slots?} At $H=96$, halving the number of slots to $K=5$ raises the MSE by 4.5\% on PEMS08 and 2.1\% on ETTm1, while doubling it to $K=20$ lowers it by only 0.9\% and 0.6\%: the default $K=10$ is close to saturation, and even $K=5$ keeps most of the gain (PEMS08: 71\% below the plain backbone; Table~\ref{tab:sensitivity}).
\end{cdraft}

\section{When does retrieval help?}
\label{sec:when-helps}
\begin{figure}[t]
\centering
\includegraphics{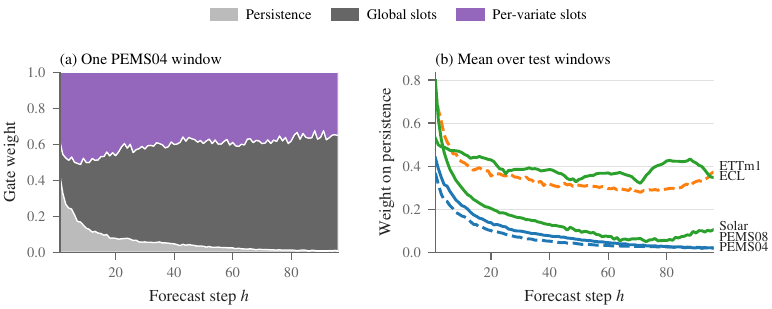}
\caption{%
\cd{The gate moves weight from persistence to retrieved futures along the horizon ($H=96$, GridTST). (a)~Gate weight per forecast step for one PEMS04 test window (median gain over the backbone; the variate with the largest target spread), by candidate group. (b)~Mean weight on persistence over 64 test windows and all variates, per dataset.}}
\label{fig:gate-example}
\end{figure}

\begin{cdraft}
The results follow one pattern: retrieval helps most where the lookback says least.

\paragraph{Further ahead, the gate trusts neighbors more.} Figure~\ref{fig:gate-example} shows that the gate puts substantial weight on persistence in the first forecast steps and moves it to retrieved futures further out: averaged over test windows, the persistence weight falls from steps 1--12 to steps 49--96 from 0.29 to 0.03 on PEMS04 and from 0.42 to 0.08 on Solar. Across all datasets, horizons and both backbones, the more weight the gate puts on neighbors, the larger the gain over the backbone (Spearman $\rho=0.71$ for GridTST, $0.53$ for iTransformer; App.~\ref{app:oracle}).

\paragraph{Longer lookbacks.} On ETTm1 ($H=96$), \ourmethod{} lowers the MSE of GridTST by 33\% at $L=48$ and 8\% at $L=96$; the gain falls to 1\% at $L=192$, and at $L=336$ \ourmethod{} is 5\% worse (Table~\ref{tab:sensitivity}). The error of \ourmethod{} changes little with the lookback (0.297--0.310), whereas the plain backbone needs a long window to reach it: retrieval substitutes for context that a short lookback lacks, but on ETTm1 it does not add to a long one. \ourmethod{} is therefore most useful where the lookback has to stay short.

\paragraph{Where retrieval fails: ETTh.} On ETTh1 and ETTh2, \ourmethod{} is worse than GridTST, and the loss is a far-horizon effect: at $H=720$ on ETTh1, \ourmethod{} is 5.5\% better over steps 1--96 and 16.5\% worse over steps 337--720. The evidence points to the level of retrieved futures drifting from the truth far ahead: a hindsight oracle over the same eligible windows shows that level-free futures would reduce the error by 26\% (ETTh1) and 36\% (ETTh2), while raw ones barely help (App.~\ref{app:oracle}).

\end{cdraft}

\section{Limitations and future work}
\label{sec:limitations}
\begin{cdraft}
\ourmethod{} needs a history of windows with observed futures, so it cannot help at a cold start. The level drift of retrieved futures limits it on non-stationary series at long horizons, and the gate cannot fully ignore irrelevant neighbors: with mismatched neighbors the model is worse than without retrieval (Sec.~\ref{sec:ablations}), and on ETTh it keeps weight on drifting futures. On ETTm1, retrieval does not improve a long lookback ($L=336$). The fixed fusion weights slightly hurt on PEMS04. Our training cache of assembled slots grows as $O(NKD(L+H))$ (36\,GB on ECL, 106\,GB on PEMS07), although the retrieval index itself is small. The main table and the ablations use one seed (three seeds in App.~\ref{app:seeds}), we evaluate at $L=96$ except for one lookback study on ETTm1, and the shared retrieval settings were not tuned per dataset. Future work includes a learned level correction of retrieved futures, better per-variate retrieval (App.~\ref{app:oracle}), approximate search, and forecasting-based anomaly detection.
\end{cdraft}

\section{Conclusion}

\begin{cdraft}
We presented \ourmethod{}, a retrieval layer that assembles neighbors both as whole past windows and per variate, and learns per forecast step and variate how much to trust them. Under a strict protocol, with a fixed lookback, causal retrieval and retrieval constants shared by all datasets, it improves two Transformer backbones in most settings of twelve benchmarks and gives the lowest MSE on eight of them. Both granularities are needed, and the gains are largest where the lookback says least: far ahead and on sensor networks. Its failures on hourly non-stationary series point to level-aware use of retrieved futures as the next step.
\end{cdraft}

\FloatBarrier %
\subsection*{AI use statement}
\begin{cdraft}
In this work, we used generative AI tools (an LLM-based coding assistant) for implementation, for running and analyzing experiments, for help with result interpretation, and for help with methodology design: the assistant proposed and screened variants of individual components, while the central idea of granular neighbor assembly and the final design are the authors'. We have not used generative AI tools for synthetic data generation, theoretical model development, mathematical formulation, proof assistance or translation, and qualitative analysis is not applicable to this work. Additionally, we used generative AI tools for figure creation, code editing, literature analysis (checking citations and related work), paper drafting and proofreading: the tools drafted parts of the text, which the authors edited and verified. We have reviewed all AI-assisted work: numbers in the text were checked against the stored result files, the causality of retrieval was audited programmatically for every retrieval bank, and every reference was checked against its source. We take responsibility for the final content of this work, including text, claims or artifacts produced with the aid of generative AI.
\end{cdraft}

\subsection*{Ethics statement}
\begin{cdraft}
This work uses public benchmark datasets, which contain no personal data. We see no specific ethical risks beyond those of forecasting in general.
\end{cdraft}

\subsection*{Reproducibility statement}
\begin{cdraft}
Sec.~\ref{sec:method} and App.~\ref{app:method-details} describe the method; App.~\ref{app:hyperparameters} lists all hyperparameters, App.~\ref{app:datasets} the datasets and splits, and App.~\ref{app:compute} the hardware and cost. The eligibility rule (Eq.~\ref{eq:eligibility}) was audited for every bank. All results use fixed seeds; App.~\ref{app:seeds} reports three seeds. Code will be released upon acceptance.
\end{cdraft}

\bibliography{bib_forecasting_paper}
\bibliographystyle{iclr2027_conference}

\clearpage
\appendix
\renewcommand{\topfraction}{0.95}
\renewcommand{\bottomfraction}{0.9}
\renewcommand{\textfraction}{0.05}
\renewcommand{\floatpagefraction}{0.75}
\setcounter{topnumber}{4}
\setcounter{bottomnumber}{3}
\setcounter{totalnumber}{6}
\let\oldsection\section
\renewcommand{\section}{\FloatBarrier\oldsection}

\section{Method details}
\label{app:method-details}
\begin{figure}[!htbp]
\centering
\resizebox{\linewidth}{!}{\definecolor{dgInk}{HTML}{222222}
\definecolor{dgMuted}{HTML}{666666}
\definecolor{dgLight}{HTML}{E0E0E0}
\definecolor{dgBlue}{HTML}{1F77B4}
\definecolor{dgViolet}{HTML}{9467BD}
\definecolor{dgAqua}{HTML}{2CA02C}
\begin{tikzpicture}[
    font=\small,
    >={Stealth[length=2mm]},
    stage/.style={draw=dgMuted, rounded corners=2pt, fill=white, align=center,
                 inner sep=3pt, line width=0.5pt, minimum height=1.15cm, text width=2.65cm},
    num/.style={circle, fill=dgInk, text=white, font=\scriptsize\bfseries, inner sep=1pt, minimum size=11pt},
    note/.style={font=\scriptsize, text=dgMuted, align=center},
    flow/.style={->, draw=dgInk, line width=0.55pt, rounded corners=4pt},
    lane/.style={font=\small, anchor=west, text=dgInk},
]
\node[lane, rotate=90, anchor=center, align=center] at (-0.95,0) {Offline\\{\scriptsize training data}};
\node[stage] (s1) at (1.1,0) {Train teacher $\phi_T$\\{\scriptsize predict the\\window's future}};
\node[stage] (s2) at (4.45,0) {Distill student $\phi$\\{\scriptsize small transformer}};
\node[stage] (s3) at (7.8,0) {Build bank\\{\scriptsize embed windows, causal ranking, assemble slots}};
\node[stage, draw=dgBlue, line width=0.8pt] (s4) at (11.15,0) {Train forecaster\\{\scriptsize gate, backbone, cross-attention}};
\foreach \a/\b in {s1/s2, s2/s3, s3/s4} {\draw[flow] (\a) -- (\b);}
\foreach \n/\k in {s1/1, s2/2, s3/3, s4/4} {\node[num] at (\n.north west) {\k};}
\node[lane, rotate=90, anchor=center, align=center] at (-0.95,-2.9) {Test period\\{\scriptsize models frozen}};
\node[stage] (s5) at (1.1,-2.9) {Add windows whose {\color{dgAqua}future} is now observed};
\node[stage] (s6) at (4.45,-2.9) {Embed query $\mathbf{X}_t$ with $\phi$};
\node[stage] (s7) at (7.8,-2.9) {Retrieve eligible\\windows, assemble\\$K$ slots};
\node[stage, draw=dgBlue, line width=0.8pt] (s8) at (11.15,-2.9) {Forecast\\$\hat{\mathbf{Y}}_t$};
\foreach \a/\b in {s5/s6, s6/s7, s7/s8} {\draw[flow] (\a) -- (\b);}
\foreach \n/\k in {s5/5, s6/6, s7/7, s8/8} {\node[num] at (\n.north west) {\k};}
\draw[flow, draw=dgMuted, densely dashed] (s2.south) -- node[note, right, pos=0.45] {$\phi$} (s6.north);
\draw[flow, draw=dgMuted, densely dashed] (s4.south) -- node[note, right, pos=0.45] {forecaster} (s8.north);
\draw[flow, draw=dgMuted, densely dashed] (s3.south) -- node[note, right, pos=0.45] {bank} (s7.north);
\draw[flow, draw=dgAqua] (s8.south) -- ++(0,-0.45) -| node[note, pos=0.25, below] {next origin $t{+}1$: newly completed windows join the bank} (s5.south);
\end{tikzpicture}}
\caption{%
\cd{Order of operations. Offline, on training data only (top): (1)~train the teacher embedder; (2)~distill the student $\phi$; (3)~build the bank: embed all windows, rank the causally eligible candidates of every query and assemble their slots; (4)~train the forecaster on these slots. Test period, with frozen models (bottom), for each forecast origin $t$: (5)~add the windows whose futures have been observed by $t$ to the bank; (6)~embed the query $\mathbf{X}_t$ with $\phi$; (7)~retrieve eligible windows and assemble $K$ slots; (8)~forecast $\hat{\mathbf{Y}}_t$. Dashed arrows: the embedder, bank and forecaster carried over from training. Green arrow: the bank grows as $t$ advances, without retraining.}}
\label{fig:pipeline}
\end{figure}
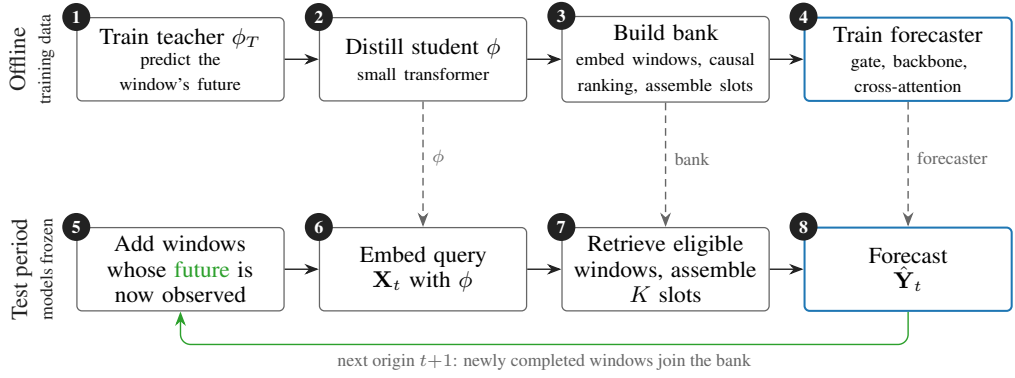

\begin{cdraft}
This section shows the three components of \ourmethod{} in detail and the order in which they are trained and used. Figure~\ref{fig:pipeline} shows the stages from training to deployment.

\paragraph{Slots.} Slot futures follow Eq.~\ref{eq:slot}; slot pasts are scaled in the same way and are used only by the gate. The per-variate match score (Sec.~\ref{sec:assembly}) compares the univariate pasts of query and candidate in the query's normalization: $0.30$ times their MSE over the full window, $0.25$ times their MSE over the last 32 steps, $0.20$ times the MSE of their first differences, $0.10$ times the squared difference of the standard deviations of their first differences, and $0.15$ times their MSE on every $(L/16)$-th step.

\begin{table}[!htbp]
\centering
\footnotesize
\begin{tabular}{@{}ll@{}}
\toprule
Symbol & Meaning \\
\midrule
$D$, $L$, $H$ & number of variates; lookback; horizon \\
$t$, $s$, $h$ & origin of the query; origin of a bank window; forecast step \\
$\mathbf{X}_t$, $\mathbf{Y}_t$, $\hat{\mathbf{Y}}_t$ & query window $\mathbf{x}_{t-L+1:t}$; target $\mathbf{x}_{t+1:t+H}$; forecast \\
$\mu_{t,d}$, $\sigma_{t,d}$ & mean and standard deviation of variate $d$ in the query window \\
\midrule
$\phi_T$, $\psi$, $\phi$ & teacher embedder; training-only future head; retrieval embedder (student) \\
$\mathbf{z}_i$, $d_z$ & teacher embedding of window $i$; its size \\
$P_i$, $Q_i$, $\tau_F$, $\tau_z$, $\lambda$ & target and embedding neighbor distributions; temperatures; KL weight \\
\midrule
$N$, $C$, $M$ & windows in the bank; fused candidates per query (200); per-variate pool (50) \\
$\rho(s)$, $\bar r_\phi$, $\bar r_j$, $w_j$ & fused rank of candidate $s$ (Eq.~\ref{eq:fusion}); normalized ranks; weight of signal $j$ \\
$w_{\mathrm{reg}}$, $w_{\mathrm{seas}}$, $w_{\mathrm{cal}}$, $d_r$ & weights of the three context signals; size of the regime features (App.~\ref{app:method-details}) \\
\midrule
$K$, $s_{k,d}$ & slots per query (5 global, 5 per-variate); source window of slot $k$ for variate $d$ \\
$\mathbf{c}_0$, $\mathbf{c}_k$ & persistence forecast; future of slot $k$, in the query's normalization (Eq.~\ref{eq:slot}) \\
$\mathbf{g}_k$ & gate weights per step and variate, summing to one over $k=0,\dots,K$ \\
$f$, $R$, $\alpha$, $\beta$ & backbone; cross-attention term; their learned scales (Eq.~\ref{eq:forecast}) \\
\bottomrule
\end{tabular}
\caption{\cd{Notation.}}
\label{tab:notation}
\end{table}

\paragraph{Retrieval embedding.} Figure~\ref{fig:embedder} shows how the embedder is trained. All reported runs add a small auxiliary term to Eq.~\ref{eq:teacher}, $\lambda\,\mathrm{KL}(P_i\,\|\,Q_i)$ with $\lambda=0.1$, which matches the embedding's neighbor distribution over a batch, $Q_i(j) = \operatorname{softmax}_{j\ne i}(\cos(\mathbf{z}_i,\mathbf{z}_j)/\tau_z)$, to one defined by future similarity, $P_i(j) = \operatorname{softmax}_{j\ne i}(-\mathrm{MSE}(\mathbf{Y}_i,\mathbf{Y}_j)/\tau_F)$, with $\tau_F=0.5$ and $\tau_z=0.1$. Removing it changes the test MSE of the full model on ETTm2 by $-0.2\%$ on average ($-0.8$ to $+1.3\%$ per horizon), so the main text presents the future-prediction objective alone. The retrieval embedder $\phi$ is a small Transformer over the $L$ time steps, distilled from the frozen teacher by minimizing $\mathrm{MSE}(\phi(\mathbf{X}),\phi_T(\mathbf{X})) + 1 - \cos(\phi(\mathbf{X}),\phi_T(\mathbf{X}))$ over training windows. The student is much smaller than the teacher, whose first layer alone has $512\,L D$ weights (42M on Traffic), so embedding every window of the bank stays cheap.

\paragraph{Rank fusion.} Eq.~\ref{eq:fusion} uses three context signals. \emph{Regime}: the Euclidean distance between fixed regime features of the two windows (a $d_r$-dimensional linear projection and binned statistics, fitted on training windows), weight $w_{\mathrm{reg}}=1$. \emph{Seasonality}: $\bar r_{\mathrm{seas}}(s)=0$ if $t-s$ is one of up to eight lags at which the autocorrelation of the training series peaks (computed on the leading regime components) and 1 otherwise, weight $w_{\mathrm{seas}}=0.25$. \emph{Calendar}: the rank of the mismatch between the time-stamp features (such as time of day and day of week) of the two windows' forecast periods, weight $w_{\mathrm{cal}}=0.5$. All ranks are normalized to $[0,1]$. The shortlist takes candidates greedily in rank order at a minimum distance from those already taken.
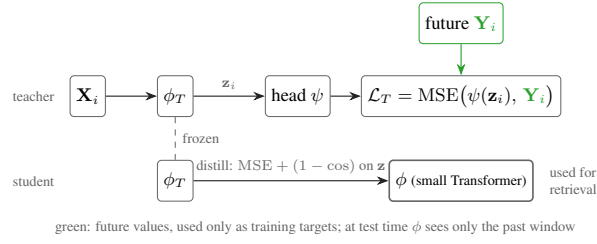
\begin{figure}[!htbp]
\centering
\resizebox{0.6\linewidth}{!}{\definecolor{dgInk}{HTML}{222222}
\definecolor{dgMuted}{HTML}{666666}
\definecolor{dgLight}{HTML}{E0E0E0}
\definecolor{dgBlue}{HTML}{1F77B4}
\definecolor{dgViolet}{HTML}{9467BD}
\definecolor{dgAqua}{HTML}{2CA02C}
\begin{tikzpicture}[
    font=\small,
    >={Stealth[length=2mm]},
    box/.style={draw=dgMuted, rounded corners=2pt, fill=white, align=center,
                inner sep=3pt, line width=0.5pt, minimum height=0.7cm},
    note/.style={font=\scriptsize, text=dgMuted, align=center},
    flow/.style={->, draw=dgInk, line width=0.55pt, rounded corners=4pt},
]
\node[box] (x) at (0,0) {$\mathbf{X}_i$};
\node[box] (phiT) at (1.5,0) {$\phi_T$};
\node[box] (psi) at (3.6,0) {head $\psi$};
\node[box] (mse) at (6.4,0) {$\mathcal{L}_T = \mathrm{MSE}\big(\psi(\mathbf{z}_i),\,{\color{dgAqua}\mathbf{Y}_i}\big)$};
\node[box, draw=dgAqua] (y) at (6.4,1.25) {future ${\color{dgAqua}\mathbf{Y}_i}$};
\draw[flow] (x) -- (phiT);
\draw[flow] (phiT) -- node[note, above] {$\mathbf{z}_i$} (psi);
\draw[flow] (psi) -- (mse);
\draw[flow, draw=dgAqua] (y) -- (mse);
\node[note, anchor=east] at (-0.45,0) {teacher};
\node[box] (teach) at (1.5,-1.45) {$\phi_T$};
\node[box, line width=0.8pt] (stud) at (6.4,-1.45) {$\phi$\ {\scriptsize(small Transformer)}};
\draw[flow] (teach) -- node[note, above] {distill: $\mathrm{MSE} + (1-\cos)$ on $\mathbf{z}$} (stud);
\draw[dashed, draw=dgMuted, line width=0.5pt] (phiT) -- node[note, right] {frozen} (teach);
\node[note, right=0.15cm of stud, align=left] {used for\\retrieval};
\node[note, anchor=east] at (-0.45,-1.45) {student};
\node[note] at (3.9,-2.25) {green: future values, used only as training targets;
at test time $\phi$ sees only the past window};
\end{tikzpicture}}
\caption{%
\cd{Training the future-aware embedder. Top: the teacher $\phi_T$ embeds a training window, and the head $\psi$ predicts the window's future from the embedding $\mathbf{z}_i$ (Eq.~\ref{eq:teacher}); green: future values, used only as training targets. Bottom: the small Transformer $\phi$ used for retrieval is distilled from the frozen teacher and sees only the past window.}}
\label{fig:embedder}
\end{figure}

\paragraph{Gate inputs.} For slot $k$ and variate $d$, the gate MLP receives: the MSE and MAE between the slot's past and the query's past, their correlation, the absolute gap at the last observed step, the variance of the slot's past, the slot type (global or per-variate), the MSE averaged over all variates (whether the past situation fits the whole system), and the calendar match between the slot's and the query's time stamps (ETT, ECL, Traffic and Weather). It outputs one logit per forecast step. Two learned per-step scales add $\tanh$ of the difference between $\mathbf{c}_k$ and the query's last value (the change the slot proposes) and $|\mathbf{c}_k - \bar{\mathbf{c}}|$ (disagreement with the mean slot future) to the logits. Persistence $\mathbf{c}_0$ has its own learned per-step logit, and a softmax over $k=0,\dots,K$ gives $\mathbf{g}_k$. The retrieval distance is not a gate input.
\end{cdraft}

\section{Implementation details and hyperparameters}
\label{app:hyperparameters}
\begin{cdraft}
Table~\ref{tab:hyperparameters-shared} lists the settings shared by all datasets, horizons and both backbones: the embedder objective, retrieval, fusion and assembly. Tables~\ref{tab:hyperparameters-gridtst} and~\ref{tab:hyperparameters-itransformer-embedder} give the per-dataset backbone and embedder settings. GridTST architectures follow the configurations of the official repository where one exists. We ran no systematic hyperparameter search. Table~\ref{tab:parameter-counts} compares parameter counts: with iTransformer, \ourmethod{} uses a smaller iTransformer than the plain baseline and, including the retrieval embedder $\phi$ (the teacher is used only in training), has fewer parameters than the plain backbone on every dataset and horizon.
\end{cdraft}
\begin{table}[!htbp]
\centering
\small
\begin{tabular}{@{}llr@{}}
\toprule
Group & Setting & Value \\
\midrule
Protocol & Lookback $L$ & 96 \\
 & Seed (single-seed runs) & 2023 \\
 & Retrieval policy & causal history index \\
 & Candidate eligibility & non-overlapping with query \\
\midrule
Embedder & Architecture & Transformer \\
 & Objective & future-MSE ranking \\
 & Future temperature $\tau_F$ & 0.5 \\
 & Embedding temperature $\tau_z$ & 0.1 \\
 & Ranking-loss weight & 0.1 \\
 & Learning rate & 0.0003 \\
 & Max epochs / patience & 50 / 10 \\
\midrule
Retrieval & Neighbors per query $K$ & 10 \\
 & Global (coherent) slots & 5 \\
 & Neighbor assembly & mixed (global + per-variate) \\
 & Global slot spacing (steps) & 1 \\
 & Per-variate slot spacing (steps) & 96 \\
 & Candidate pool & 50 \\
 & Shortlist policy & episode-diverse \\
 & Ranking & hybrid rank fusion \\
 & Ranked candidates (hybrid) & 200 \\
 & Regime weight & 1 \\
 & Seasonal weight & 0.25 \\
 & Calendar weight & 0.5 \\
 & Neighbor future transport & raw \\
\midrule
Forecaster & Max epochs / patience & 10 / 10 \\
\bottomrule
\end{tabular}
\caption{\papercaption{tab:hyperparameters-shared}}
\label{tab:hyperparameters-shared}
\end{table}

\begin{table}[!htbp]
\centering
\scriptsize
\setlength{\tabcolsep}{4pt}
\begin{tabular}{@{}lcccccc@{}}
\toprule
 & \multicolumn{6}{c}{GridTST backbone and training} \\
\cmidrule(lr){2-7}
Dataset & batch & lr & $d_\mathrm{model}$ & layers & patch & stride \\
\midrule
PEMS03 & 32 & 0.001 & 64 & 3 & 16 & 16 \\
PEMS04 & 32 & 0.0005 & 64 & 3 & 16 & 16 \\
PEMS07 & 32 & 0.001 & 64 & 3 & 16 & 16 \\
PEMS08 & 32 & 0.001 & 64 & 3 & 16 & 16 \\
\midrule
ETTh1 & 32 & 0.001 & 96/96/128/256 & 3 & 48/48/48/64 & 48/48/96/16 \\
ETTh2 & 32 & 0.001/0.001/0.001/0.0001 & 96/96/128/256 & 3 & 48/48/48/64 & 48/48/96/16 \\
ETTm1 & 32 & 0.001 & 64/64/32/64 & 3 & 48/48/32/4 & 48/48/16/4 \\
ETTm2 & 32 & 0.001 & 64/64/32/64 & 3 & 48/48/32/4 & 48/48/16/4 \\
ECL & 32 & 0.001 & 256 & 3 & 8 & 8 \\
Solar & 32 & 0.001 & 64 & 3 & 16 & 16 \\
Traffic & 8 & 0.001 & 256 & 5 & 16 & 8 \\
Weather & 32 & 0.001 & 128 & 4 & 16 & 16 \\
\bottomrule
\end{tabular}
\caption{\papercaption{tab:hyperparameters-gridtst}}
\label{tab:hyperparameters-gridtst}
\end{table}

\begin{table}[!htbp]
\centering
\scriptsize
\setlength{\tabcolsep}{4pt}
\begin{tabular}{@{}lccccccccc@{}}
\toprule
 & \multicolumn{4}{c}{iTransformer} & \multicolumn{5}{c}{Retrieval embedder} \\
\cmidrule(lr){2-5}\cmidrule(lr){6-10}
Dataset & batch & lr & $d_\mathrm{model}$ & layers & $d_\mathrm{model}$ & emb. & layers & heads & cal. \\
\midrule
PEMS03 & 32 & 0.001 & 384 & 4 & 256 & 128 & 2 & 4 & no \\
PEMS04 & 32 & 0.0005 & 896 & 4 & 256 & 128 & 2 & 4 & no \\
PEMS07 & 32 & 0.001 & 64/64/256/256 & 3/3/4/4 & 256 & 128 & 2 & 4 & no \\
PEMS08 & 32 & 0.001 & 64/64/256/256 & 3/3/4/4 & 256 & 128 & 2 & 4 & no \\
\midrule
ETTh1 & 32 & 0.001 & 64 & 1 & 32 & 32 & 1 & 2 & yes \\
ETTh2 & 32 & 0.001/0.001/0.001/0.0001 & 64 & 1 & 32 & 32 & 1 & 2 & yes \\
ETTm1 & 32 & 0.001 & 64 & 1 & 32 & 32 & 1 & 2 & yes \\
ETTm2 & 32 & 0.001 & 64 & 1 & 32 & 32 & 1 & 2 & yes \\
ECL & 32 & 0.001 & 350 & 2 & 256 & 128 & 2 & 4 & yes \\
Solar & 32 & 0.001 & 200 & 4 & 256 & 128 & 2 & 4 & no \\
Traffic & 8 & 0.001 & 300 & 4 & 256 & 128 & 2 & 4 & yes \\
Weather & 32 & 0.001 & 300 & 2 & 32 & 128 & 1 & 2 & yes \\
\bottomrule
\end{tabular}
\caption{\papercaption{tab:hyperparameters-itransformer-embedder}}
\label{tab:hyperparameters-itransformer-embedder}
\end{table}

\begin{table}[!htbp]
\centering
\small
\setlength{\tabcolsep}{4pt}
\begin{tabular}{@{}lr@{}c@{}lr@{}c@{}lr@{}c@{}lr@{}c@{}lr@{}c@{}lr@{}c@{}lr@{}c@{}l@{}}
\toprule
 & \multicolumn{9}{c}{GridTST} & \multicolumn{9}{c}{iTransformer} & \\
\cmidrule(lr){2-10}\cmidrule(lr){11-19}
Dataset & \multicolumn{3}{c}{Backbone} & \multicolumn{3}{c}{GNA} & \multicolumn{3}{c}{Ratio} & \multicolumn{3}{c}{Backbone} & \multicolumn{3}{c}{GNA} & \multicolumn{3}{c}{Ratio} & \multicolumn{3}{c}{Embedder} \\
\midrule
PEMS03 & 0.16 & \,--\, & 0.20 & 2.59 & \,--\, & 2.70 & 13.84 & \,--\, & 16.46 & 6.37 & \,--\, & 6.41 & 6.01 & \,--\, & 6.11 & 0.94 & \,--\, & 0.95 & 1.81 & &  \\
PEMS04 & 0.16 & \,--\, & 0.20 & 5.25 & \,--\, & 5.45 & 27.90 & \,--\, & 33.34 & 25.32 & \,--\, & 25.41 & 24.44 & \,--\, & 24.67 & 0.97 & &  & 1.79 & &  \\
PEMS07 & 0.16 & \,--\, & 0.20 & 2.14 & \,--\, & 2.49 & 12.69 & \,--\, & 13.98 & 3.21 & \,--\, & 6.41 & 2.06 & \,--\, & 3.91 & 0.61 & \,--\, & 0.64 & 1.96 & &  \\
PEMS08 & 0.16 & \,--\, & 0.19 & 1.93 & \,--\, & 2.29 & 11.73 & \,--\, & 12.90 & 3.21 & \,--\, & 6.41 & 1.85 & \,--\, & 3.71 & 0.57 & \,--\, & 0.58 & 1.75 & &  \\
\midrule
ETTh1 & 0.29 & \,--\, & 1.94 & 0.36 & \,--\, & 2.12 & 1.09 & \,--\, & 1.25 & 0.84 & \,--\, & 3.58 & 0.10 & \,--\, & 0.24 & 0.05 & \,--\, & 0.14 & 0.02 & &  \\
ETTh2 & 0.29 & \,--\, & 1.94 & 0.36 & \,--\, & 2.12 & 1.09 & \,--\, & 1.25 & 0.22 & \,--\, & 0.30 & 0.10 & \,--\, & 0.24 & 0.43 & \,--\, & 0.79 & 0.02 & &  \\
ETTm1 & 0.13 & \,--\, & 1.30 & 0.23 & \,--\, & 1.47 & 1.13 & \,--\, & 1.78 & 0.22 & \,--\, & 0.30 & 0.10 & \,--\, & 0.24 & 0.43 & \,--\, & 0.79 & 0.02 & &  \\
ETTm2 & 0.13 & \,--\, & 1.30 & 0.23 & \,--\, & 1.47 & 1.13 & \,--\, & 1.78 & 0.22 & \,--\, & 0.30 & 0.10 & \,--\, & 0.24 & 0.43 & \,--\, & 0.79 & 0.02 & &  \\
ECL & 1.51 & \,--\, & 3.59 & 4.00 & \,--\, & 6.60 & 1.84 & \,--\, & 2.64 & 4.83 & \,--\, & 5.15 & 3.92 & \,--\, & 4.66 & 0.81 & \,--\, & 0.91 & 1.80 & &  \\
Solar & 0.19 & \,--\, & 0.48 & 2.16 & \,--\, & 2.72 & 5.73 & \,--\, & 11.07 & 3.26 & \,--\, & 3.58 & 2.96 & \,--\, & 3.37 & 0.91 & \,--\, & 0.94 & 1.74 & &  \\
Traffic & 2.28 & \,--\, & 4.20 & 4.77 & \,--\, & 7.15 & 1.70 & \,--\, & 2.09 & 6.41 & \,--\, & 6.73 & 4.62 & \,--\, & 5.27 & 0.72 & \,--\, & 0.78 & 1.95 & &  \\
Weather & 0.62 & \,--\, & 1.18 & 1.14 & \,--\, & 2.16 & 1.83 & \,--\, & 1.84 & 4.83 & \,--\, & 5.15 & 1.62 & \,--\, & 2.26 & 0.33 & \,--\, & 0.44 & 0.02 & &  \\
\bottomrule
\end{tabular}
\caption{\papercaption{tab:parameter-counts}}
\label{tab:parameter-counts}
\end{table}

\section{Datasets}
\label{app:datasets}
\begin{cdraft}
Table~\ref{tab:datasets} summarizes the datasets. We use the standard chronological splits and standardize every variate with the mean and standard deviation of the training split; all errors are computed on this scale.
\end{cdraft}
\begin{table}[!htbp]
\centering
\small
\setlength{\tabcolsep}{3pt}
\begin{tabular}{@{}lrr@{\,}lrrrcrr@{\hspace{3pt}}r@{\hspace{3pt}}r@{\hspace{3pt}}r@{}}
\toprule
Dataset & Variates & \multicolumn{2}{c}{Sampling} & Train & Val & Test & Split & Lookback & \multicolumn{4}{c}{Horizons} \\
\midrule
PEMS03 & 358 & 5 & min & 15,724 & 5,242 & 5,242 & 60:20:20 & 96 & \{12, & 24, & 48, & 96\} \\
PEMS04 & 307 & 5 & min & 10,195 & 3,398 & 3,399 & 60:20:20 & 96 & \{12, & 24, & 48, & 96\} \\
PEMS07 & 883 & 5 & min & 16,934 & 5,645 & 5,645 & 60:20:20 & 96 & \{12, & 24, & 48, & 96\} \\
PEMS08 & 170 & 5 & min & 10,713 & 3,571 & 3,572 & 60:20:20 & 96 & \{12, & 24, & 48, & 96\} \\
\midrule
ETTh1/ETTh2 & 7 & 1 & h & 8,640 & 2,880 & 2,880 & 60:20:20 & 96 & \{96, & 192, & 336, & 720\} \\
ETTm1/ETTm2 & 7 & 15 & min & 34,560 & 11,520 & 11,520 & 60:20:20 & 96 & \{96, & 192, & 336, & 720\} \\
ECL & 321 & 1 & h & 18,412 & 2,632 & 5,260 & 70:10:20 & 96 & \{96, & 192, & 336, & 720\} \\
Solar & 137 & 10 & min & 36,792 & 5,256 & 10,512 & 70:10:20 & 96 & \{96, & 192, & 336, & 720\} \\
Traffic & 862 & 1 & h & 12,280 & 1,756 & 3,508 & 70:10:20 & 96 & \{96, & 192, & 336, & 720\} \\
Weather & 21 & 10 & min & 36,887 & 5,270 & 10,539 & 70:10:20 & 96 & \{96, & 192, & 336, & 720\} \\
\bottomrule
\end{tabular}
\caption{\papercaption{tab:datasets}}
\label{tab:datasets}
\end{table}

\section{Full results}
\label{app:full-results}
\begin{figure}[!htbp]
\centering
\includegraphics{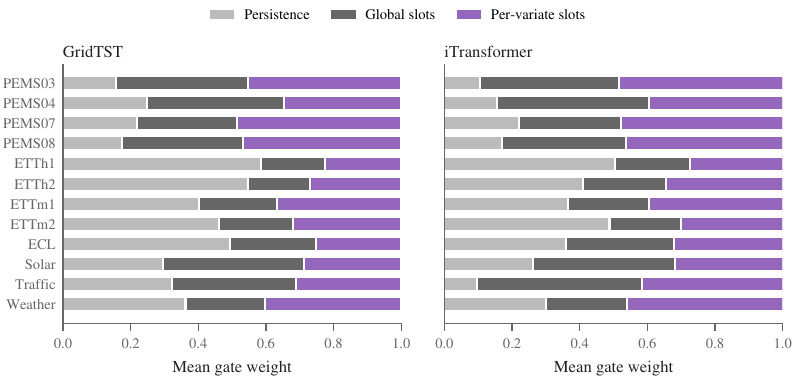}
\caption{%
\cd{How the gate splits its weight between persistence ($\mathbf{c}_0$), global slots and per-variate slots, averaged over forecast steps, test windows and variates. The backbone term $\alpha f(\mathbf{X}_t)$ is added outside the gate and is not part of this split.}}
\label{fig:gate-composition}
\end{figure}

\begin{cdraft}
Table~\ref{tab:retrieval-vs-baseline} compares each backbone with and without retrieval. The per-horizon results and the comparison with all 18 baselines are in Tables~\ref{tab:appendix_per_horizon_table} and~\ref{tab:appendix_all_models_table}. Figure~\ref{fig:mse-by-horizon} shows the MSE per horizon, and Fig.~\ref{fig:gate-composition} how the gate distributes its weight on average.
\end{cdraft}

\begin{table}[!htbp]
\centering
\footnotesize
\setlength{\tabcolsep}{4pt}
\begin{tabular}{@{}lcccccccc@{}}
\toprule
& \multicolumn{4}{c}{Backbone only} & \multicolumn{4}{c}{GNA (this work)} \\
\cmidrule(lr){2-5}\cmidrule(l){6-9}
Dataset & \multicolumn{2}{c}{iTransformer} & \multicolumn{2}{c}{GridTST} & \multicolumn{2}{c}{iTransformer} & \multicolumn{2}{c}{GridTST} \\
& \multicolumn{2}{c}{\tiny ICLR \citeyear{liuITransformerInvertedTransformers2024}, our run} & \multicolumn{2}{c}{\tiny arXiv \citeyear{GridTST}, our run} & \multicolumn{2}{c}{\tiny this work} & \multicolumn{2}{c}{\tiny this work} \\
\cmidrule(lr){2-3}\cmidrule(lr){4-5}\cmidrule(lr){6-7}\cmidrule(l){8-9}
& MSE & MAE & MSE & MAE & MSE & MAE & MSE & MAE \\
\midrule
\multicolumn{9}{@{}l}{\textit{Short-term forecasting}} \\
\addlinespace[2pt]
PEMS03 & 0.119 & 0.227 & 0.153 & 0.260 & \underline{\textcolor{blue}{0.084}} & \underline{\textcolor{blue}{0.183}} & \textbf{\textcolor{red}{0.082}} & \textbf{\textcolor{red}{0.182}} \\
PEMS04 & 0.120 & 0.231 & 0.174 & 0.283 & \underline{\textcolor{blue}{0.083}} & \underline{\textcolor{blue}{0.184}} & \textbf{\textcolor{red}{0.080}} & \textbf{\textcolor{red}{0.182}} \\
PEMS07 & 0.581 & 0.503 & 0.134 & 0.241 & \underline{\textcolor{blue}{0.067}} & \underline{\textcolor{blue}{0.156}} & \textbf{\textcolor{red}{0.065}} & \textbf{\textcolor{red}{0.155}} \\
PEMS08 & 0.259 & 0.312 & \underline{\textcolor{blue}{0.184}} & \underline{\textcolor{blue}{0.275}} & \textbf{\textcolor{red}{0.076}} & \textbf{\textcolor{red}{0.176}} & \textbf{\textcolor{red}{0.076}} & \textbf{\textcolor{red}{0.176}} \\
\addlinespace[5pt]
\multicolumn{9}{@{}l}{\textit{Long-term forecasting}} \\
\addlinespace[2pt]
ETTh1 & \underline{\textcolor{blue}{0.457}} & \underline{\textcolor{blue}{0.449}} & \textbf{\textcolor{red}{0.456}} & \textbf{\textcolor{red}{0.448}} & 0.473 & 0.464 & 0.463 & 0.459 \\
ETTh2 & \textbf{\textcolor{red}{0.384}} & \textbf{\textcolor{red}{0.407}} & \underline{\textcolor{blue}{0.395}} & \underline{\textcolor{blue}{0.412}} & 0.427 & 0.441 & 0.420 & 0.437 \\
ETTm1 & 0.408 & 0.412 & 0.386 & 0.402 & \underline{\textcolor{blue}{0.365}} & \underline{\textcolor{blue}{0.389}} & \textbf{\textcolor{red}{0.359}} & \textbf{\textcolor{red}{0.386}} \\
ETTm2 & 0.292 & 0.336 & 0.287 & 0.332 & \underline{\textcolor{blue}{0.278}} & \underline{\textcolor{blue}{0.331}} & \textbf{\textcolor{red}{0.272}} & \textbf{\textcolor{red}{0.329}} \\
ECL & 0.176 & 0.267 & 0.174 & 0.266 & \underline{\textcolor{blue}{0.154}} & \underline{\textcolor{blue}{0.255}} & \textbf{\textcolor{red}{0.153}} & \textbf{\textcolor{red}{0.252}} \\
Solar & 0.236 & 0.262 & 0.228 & 0.278 & \underline{\textcolor{blue}{0.201}} & \textbf{\textcolor{red}{0.243}} & \textbf{\textcolor{red}{0.198}} & \underline{\textcolor{blue}{0.245}} \\
Traffic & 0.423 & 0.283 & 0.412 & 0.271 & \underline{\textcolor{blue}{0.409}} & \textbf{\textcolor{red}{0.263}} & \textbf{\textcolor{red}{0.397}} & \underline{\textcolor{blue}{0.269}} \\
Weather & 0.260 & 0.280 & 0.245 & \underline{\textcolor{blue}{0.272}} & \underline{\textcolor{blue}{0.239}} & 0.273 & \textbf{\textcolor{red}{0.235}} & \textbf{\textcolor{red}{0.270}} \\
\midrule
1st count & 1 & 1 & 1 & 1 & 1 & 3 & \textbf{\textcolor{red}{10}} & \textbf{\textcolor{red}{8}} \\
2nd count & 1 & 1 & 2 & 3 & \underline{\textcolor{blue}{9}} & \underline{\textcolor{blue}{6}} & 0 & 2 \\
Retrieval vs baseline & 2 & 2 & 2 & 2 & \textbf{10} & \textbf{10} & \textbf{10} & \textbf{10} \\
\midrule
\multicolumn{9}{@{}l}{\textit{Per-task}} \\
1st count & 4 & 5 & 0 & 6 & 14 & 18 & \textbf{\textcolor{red}{39}} & \textbf{\textcolor{red}{23}} \\
2nd count & 4 & 4 & 12 & 7 & \underline{\textcolor{blue}{28}} & \underline{\textcolor{blue}{19}} & 4 & 18 \\
Retrieval vs baseline & 5 & 8 & 5 & 10 & \textbf{42} & \textbf{40} & \textbf{43} & \textbf{37} \\
\midrule
\multicolumn{9}{@{}l}{\footnotesize \textbf{\textcolor{red}{Red, bold}}: best in row.} \\
\multicolumn{9}{@{}l}{\footnotesize \underline{\textcolor{blue}{Blue, underline}}: 2nd-best in row.} \\
\bottomrule
\end{tabular}

\caption{%
\cd{Retrieval vs.\ no retrieval with the same backbone, data and lookback ($L=96$): test MSE and MAE averaged over the four horizons, for GridTST and iTransformer without retrieval (backbone only) and with retrieval (\ourmethod{}). All numbers are our runs. With iTransformer, \ourmethod{} uses a smaller iTransformer configuration than the plain baseline (App.~\ref{app:hyperparameters}). Our iTransformer runs are far worse than the published results on PEMS07 and PEMS08, a known reproduction issue; Table~\ref{tab:main_comparison_table} uses the published numbers.}}
\label{tab:retrieval-vs-baseline}
\end{table}
\providecommand{\ourmethod}{GNA}
\begin{table}[!htbp]
\centering
\scriptsize
\setlength{\tabcolsep}{3pt}
\begin{tabular}{llcccccccc}
\toprule
 & & \multicolumn{4}{c}{\textit{\ourmethod{} (this work)}} & \multicolumn{4}{c}{\textit{Baselines}} \\
\cmidrule(lr){3-6}\cmidrule(lr){7-10}
Dataset & $H$ & \multicolumn{2}{c}{w/ GridTST} & \multicolumn{2}{c}{w/ iTransformer} & \multicolumn{2}{c}{GridTST} & \multicolumn{2}{c}{iTransformer} \\
 & & \multicolumn{2}{c}{\tiny } & \multicolumn{2}{c}{\tiny } & \multicolumn{2}{c}{\tiny arXiv \citeyear{GridTST}, our run} & \multicolumn{2}{c}{\tiny ICLR \citeyear{liuITransformerInvertedTransformers2024}, our run} \\
\cmidrule(lr){3-4}\cmidrule(lr){5-6}\cmidrule(lr){7-8}\cmidrule(lr){9-10}
 & & MSE & MAE & MSE & MAE & MSE & MAE & MSE & MAE \\
\midrule
PEMS03 & 12 & \textbf{\textcolor{red}{0.057}} & \textbf{\textcolor{red}{0.156}} & \underline{\textcolor{blue}{0.060}} & \underline{\textcolor{blue}{0.160}} & 0.069 & 0.177 & 0.072 & 0.177 \\
 & 24 & \textbf{\textcolor{red}{0.070}} & \textbf{\textcolor{red}{0.171}} & \underline{\textcolor{blue}{0.072}} & \underline{\textcolor{blue}{0.174}} & 0.099 & 0.215 & 0.094 & 0.203 \\
 & 48 & \textbf{\textcolor{red}{0.092}} & \underline{\textcolor{blue}{0.193}} & \underline{\textcolor{blue}{0.093}} & \textbf{\textcolor{red}{0.192}} & 0.173 & 0.283 & 0.132 & 0.243 \\
 & 96 & \textbf{\textcolor{red}{0.111}} & \underline{\textcolor{blue}{0.209}} & \textbf{\textcolor{red}{0.111}} & \textbf{\textcolor{red}{0.207}} & 0.273 & 0.366 & \underline{\textcolor{blue}{0.176}} & 0.285 \\
 & \textit{Avg} & \textbf{\textcolor{red}{0.082}} & \textbf{\textcolor{red}{0.182}} & \underline{\textcolor{blue}{0.084}} & \underline{\textcolor{blue}{0.183}} & 0.153 & 0.260 & 0.119 & 0.227 \\
\midrule
PEMS04 & 12 & \textbf{\textcolor{red}{0.066}} & \textbf{\textcolor{red}{0.165}} & \underline{\textcolor{blue}{0.070}} & \underline{\textcolor{blue}{0.171}} & 0.086 & 0.194 & 0.081 & 0.188 \\
 & 24 & \textbf{\textcolor{red}{0.075}} & \textbf{\textcolor{red}{0.176}} & \underline{\textcolor{blue}{0.077}} & \underline{\textcolor{blue}{0.178}} & 0.121 & 0.240 & 0.100 & 0.212 \\
 & 48 & \textbf{\textcolor{red}{0.085}} & \textbf{\textcolor{red}{0.187}} & \underline{\textcolor{blue}{0.086}} & \underline{\textcolor{blue}{0.188}} & 0.189 & 0.303 & 0.133 & 0.246 \\
 & 96 & \textbf{\textcolor{red}{0.096}} & \textbf{\textcolor{red}{0.199}} & \textbf{\textcolor{red}{0.096}} & \textbf{\textcolor{red}{0.199}} & 0.300 & 0.395 & \underline{\textcolor{blue}{0.165}} & \underline{\textcolor{blue}{0.277}} \\
 & \textit{Avg} & \textbf{\textcolor{red}{0.080}} & \textbf{\textcolor{red}{0.182}} & \underline{\textcolor{blue}{0.083}} & \underline{\textcolor{blue}{0.184}} & 0.174 & 0.283 & 0.120 & 0.231 \\
\midrule
PEMS07 & 12 & \textbf{\textcolor{red}{0.050}} & \textbf{\textcolor{red}{0.140}} & \underline{\textcolor{blue}{0.051}} & \underline{\textcolor{blue}{0.141}} & 0.064 & 0.164 & 0.066 & 0.164 \\
 & 24 & \textbf{\textcolor{red}{0.060}} & \textbf{\textcolor{red}{0.150}} & \textbf{\textcolor{red}{0.060}} & \underline{\textcolor{blue}{0.152}} & 0.096 & 0.201 & \underline{\textcolor{blue}{0.087}} & 0.190 \\
 & 48 & \textbf{\textcolor{red}{0.070}} & \textbf{\textcolor{red}{0.161}} & \underline{\textcolor{blue}{0.073}} & \underline{\textcolor{blue}{0.162}} & 0.153 & 0.265 & 0.689 & 0.619 \\
 & 96 & \textbf{\textcolor{red}{0.081}} & \textbf{\textcolor{red}{0.168}} & \underline{\textcolor{blue}{0.083}} & \underline{\textcolor{blue}{0.169}} & 0.224 & 0.335 & 1.482 & 1.037 \\
 & \textit{Avg} & \textbf{\textcolor{red}{0.065}} & \textbf{\textcolor{red}{0.155}} & \underline{\textcolor{blue}{0.067}} & \underline{\textcolor{blue}{0.156}} & 0.134 & 0.241 & 0.581 & 0.503 \\
\midrule
PEMS08 & 12 & \textbf{\textcolor{red}{0.059}} & \underline{\textcolor{blue}{0.157}} & \textbf{\textcolor{red}{0.059}} & \textbf{\textcolor{red}{0.156}} & \underline{\textcolor{blue}{0.078}} & 0.182 & 0.088 & 0.193 \\
 & 24 & \textbf{\textcolor{red}{0.068}} & \textbf{\textcolor{red}{0.169}} & \textbf{\textcolor{red}{0.068}} & \textbf{\textcolor{red}{0.169}} & \underline{\textcolor{blue}{0.114}} & \underline{\textcolor{blue}{0.223}} & 0.138 & 0.243 \\
 & 48 & \textbf{\textcolor{red}{0.079}} & \textbf{\textcolor{red}{0.181}} & \underline{\textcolor{blue}{0.080}} & \underline{\textcolor{blue}{0.182}} & 0.188 & 0.290 & 0.319 & 0.353 \\
 & 96 & \textbf{\textcolor{red}{0.097}} & \underline{\textcolor{blue}{0.198}} & \textbf{\textcolor{red}{0.097}} & \textbf{\textcolor{red}{0.196}} & \underline{\textcolor{blue}{0.354}} & 0.406 & 0.492 & 0.459 \\
 & \textit{Avg} & \textbf{\textcolor{red}{0.076}} & \textbf{\textcolor{red}{0.176}} & \textbf{\textcolor{red}{0.076}} & \textbf{\textcolor{red}{0.176}} & \underline{\textcolor{blue}{0.184}} & \underline{\textcolor{blue}{0.275}} & 0.259 & 0.312 \\
\midrule
ETTh1 & 96 & \textbf{\textcolor{red}{0.378}} & \underline{\textcolor{blue}{0.402}} & 0.384 & 0.404 & \underline{\textcolor{blue}{0.379}} & \textbf{\textcolor{red}{0.401}} & 0.387 & 0.405 \\
 & 192 & 0.446 & 0.447 & \textbf{\textcolor{red}{0.435}} & 0.438 & \underline{\textcolor{blue}{0.436}} & \textbf{\textcolor{red}{0.433}} & 0.441 & \underline{\textcolor{blue}{0.436}} \\
 & 336 & \textbf{\textcolor{red}{0.461}} & \textbf{\textcolor{red}{0.457}} & \underline{\textcolor{blue}{0.473}} & 0.465 & 0.490 & 0.464 & 0.491 & \underline{\textcolor{blue}{0.462}} \\
 & 720 & 0.566 & 0.529 & 0.599 & 0.549 & \underline{\textcolor{blue}{0.521}} & \textbf{\textcolor{red}{0.493}} & \textbf{\textcolor{red}{0.509}} & \underline{\textcolor{blue}{0.494}} \\
 & \textit{Avg} & 0.463 & 0.459 & 0.473 & 0.464 & \textbf{\textcolor{red}{0.456}} & \textbf{\textcolor{red}{0.448}} & \underline{\textcolor{blue}{0.457}} & \underline{\textcolor{blue}{0.449}} \\
\midrule
ETTh2 & 96 & \underline{\textcolor{blue}{0.305}} & 0.358 & \textbf{\textcolor{red}{0.301}} & 0.357 & 0.318 & \underline{\textcolor{blue}{0.356}} & \textbf{\textcolor{red}{0.301}} & \textbf{\textcolor{red}{0.350}} \\
 & 192 & 0.386 & 0.408 & 0.390 & 0.414 & \underline{\textcolor{blue}{0.383}} & \underline{\textcolor{blue}{0.401}} & \textbf{\textcolor{red}{0.380}} & \textbf{\textcolor{red}{0.399}} \\
 & 336 & 0.440 & 0.451 & \textbf{\textcolor{red}{0.420}} & 0.444 & 0.426 & \underline{\textcolor{blue}{0.433}} & \underline{\textcolor{blue}{0.424}} & \textbf{\textcolor{red}{0.432}} \\
 & 720 & 0.551 & 0.529 & 0.595 & 0.551 & \underline{\textcolor{blue}{0.451}} & \underline{\textcolor{blue}{0.458}} & \textbf{\textcolor{red}{0.430}} & \textbf{\textcolor{red}{0.447}} \\
 & \textit{Avg} & 0.420 & 0.437 & 0.427 & 0.441 & \underline{\textcolor{blue}{0.395}} & \underline{\textcolor{blue}{0.412}} & \textbf{\textcolor{red}{0.384}} & \textbf{\textcolor{red}{0.407}} \\
\midrule
ETTm1 & 96 & \textbf{\textcolor{red}{0.297}} & \textbf{\textcolor{red}{0.341}} & \underline{\textcolor{blue}{0.298}} & \textbf{\textcolor{red}{0.341}} & 0.323 & \underline{\textcolor{blue}{0.363}} & 0.342 & 0.377 \\
 & 192 & \textbf{\textcolor{red}{0.341}} & \underline{\textcolor{blue}{0.371}} & \underline{\textcolor{blue}{0.342}} & \textbf{\textcolor{red}{0.370}} & 0.360 & 0.380 & 0.383 & 0.396 \\
 & 336 & \textbf{\textcolor{red}{0.373}} & \underline{\textcolor{blue}{0.400}} & \textbf{\textcolor{red}{0.373}} & \textbf{\textcolor{red}{0.398}} & \underline{\textcolor{blue}{0.401}} & 0.412 & 0.418 & 0.418 \\
 & 720 & \textbf{\textcolor{red}{0.424}} & \textbf{\textcolor{red}{0.431}} & \underline{\textcolor{blue}{0.448}} & \underline{\textcolor{blue}{0.446}} & 0.463 & 0.453 & 0.487 & 0.457 \\
 & \textit{Avg} & \textbf{\textcolor{red}{0.359}} & \textbf{\textcolor{red}{0.386}} & \underline{\textcolor{blue}{0.365}} & \underline{\textcolor{blue}{0.389}} & 0.386 & 0.402 & 0.408 & 0.412 \\
\midrule
ETTm2 & 96 & \textbf{\textcolor{red}{0.169}} & \textbf{\textcolor{red}{0.258}} & \underline{\textcolor{blue}{0.172}} & \underline{\textcolor{blue}{0.261}} & 0.179 & 0.262 & 0.186 & 0.272 \\
 & 192 & \textbf{\textcolor{red}{0.223}} & \textbf{\textcolor{red}{0.298}} & \underline{\textcolor{blue}{0.234}} & \underline{\textcolor{blue}{0.305}} & 0.246 & 0.306 & 0.254 & 0.314 \\
 & 336 & \underline{\textcolor{blue}{0.297}} & \underline{\textcolor{blue}{0.346}} & \textbf{\textcolor{red}{0.290}} & \textbf{\textcolor{red}{0.342}} & 0.310 & 0.348 & 0.316 & 0.351 \\
 & 720 & \textbf{\textcolor{red}{0.398}} & 0.414 & 0.416 & 0.415 & \underline{\textcolor{blue}{0.413}} & \underline{\textcolor{blue}{0.411}} & 0.414 & \textbf{\textcolor{red}{0.407}} \\
 & \textit{Avg} & \textbf{\textcolor{red}{0.272}} & \textbf{\textcolor{red}{0.329}} & \underline{\textcolor{blue}{0.278}} & \underline{\textcolor{blue}{0.331}} & 0.287 & 0.332 & 0.292 & 0.336 \\
\midrule
ECL & 96 & \underline{\textcolor{blue}{0.125}} & \underline{\textcolor{blue}{0.223}} & \textbf{\textcolor{red}{0.124}} & \textbf{\textcolor{red}{0.222}} & 0.149 & 0.240 & 0.148 & 0.239 \\
 & 192 & \textbf{\textcolor{red}{0.144}} & \textbf{\textcolor{red}{0.241}} & \textbf{\textcolor{red}{0.144}} & \underline{\textcolor{blue}{0.242}} & \underline{\textcolor{blue}{0.162}} & 0.254 & 0.166 & 0.258 \\
 & 336 & \underline{\textcolor{blue}{0.158}} & \underline{\textcolor{blue}{0.257}} & \textbf{\textcolor{red}{0.155}} & \textbf{\textcolor{red}{0.256}} & 0.176 & 0.270 & 0.179 & 0.273 \\
 & 720 & \textbf{\textcolor{red}{0.186}} & \textbf{\textcolor{red}{0.286}} & \underline{\textcolor{blue}{0.193}} & \underline{\textcolor{blue}{0.298}} & 0.209 & 0.300 & 0.211 & 0.300 \\
 & \textit{Avg} & \textbf{\textcolor{red}{0.153}} & \textbf{\textcolor{red}{0.252}} & \underline{\textcolor{blue}{0.154}} & \underline{\textcolor{blue}{0.255}} & 0.174 & 0.266 & 0.176 & 0.267 \\
\midrule
Solar & 96 & \textbf{\textcolor{red}{0.183}} & \underline{\textcolor{blue}{0.217}} & \underline{\textcolor{blue}{0.184}} & \textbf{\textcolor{red}{0.214}} & 0.196 & 0.253 & 0.206 & 0.236 \\
 & 192 & \textbf{\textcolor{red}{0.190}} & \underline{\textcolor{blue}{0.236}} & \underline{\textcolor{blue}{0.197}} & \textbf{\textcolor{red}{0.232}} & 0.223 & 0.272 & 0.239 & 0.263 \\
 & 336 & \textbf{\textcolor{red}{0.201}} & \textbf{\textcolor{red}{0.258}} & \underline{\textcolor{blue}{0.205}} & \underline{\textcolor{blue}{0.259}} & 0.239 & 0.289 & 0.251 & 0.274 \\
 & 720 & \textbf{\textcolor{red}{0.217}} & \underline{\textcolor{blue}{0.269}} & \underline{\textcolor{blue}{0.218}} & \textbf{\textcolor{red}{0.268}} & 0.255 & 0.299 & 0.250 & 0.275 \\
 & \textit{Avg} & \textbf{\textcolor{red}{0.198}} & \underline{\textcolor{blue}{0.245}} & \underline{\textcolor{blue}{0.201}} & \textbf{\textcolor{red}{0.243}} & 0.228 & 0.278 & 0.236 & 0.262 \\
\midrule
Traffic & 96 & \textbf{\textcolor{red}{0.365}} & \underline{\textcolor{blue}{0.250}} & \underline{\textcolor{blue}{0.370}} & \textbf{\textcolor{red}{0.242}} & 0.388 & 0.260 & 0.394 & 0.270 \\
 & 192 & \textbf{\textcolor{red}{0.391}} & \underline{\textcolor{blue}{0.264}} & \underline{\textcolor{blue}{0.393}} & \textbf{\textcolor{red}{0.254}} & 0.403 & 0.265 & 0.413 & 0.277 \\
 & 336 & \textbf{\textcolor{red}{0.395}} & \underline{\textcolor{blue}{0.268}} & \underline{\textcolor{blue}{0.405}} & \textbf{\textcolor{red}{0.262}} & 0.414 & 0.271 & 0.425 & 0.283 \\
 & 720 & \textbf{\textcolor{red}{0.437}} & \underline{\textcolor{blue}{0.293}} & 0.468 & 0.296 & \underline{\textcolor{blue}{0.443}} & \textbf{\textcolor{red}{0.288}} & 0.460 & 0.301 \\
 & \textit{Avg} & \textbf{\textcolor{red}{0.397}} & \underline{\textcolor{blue}{0.269}} & \underline{\textcolor{blue}{0.409}} & \textbf{\textcolor{red}{0.263}} & 0.412 & 0.271 & 0.423 & 0.283 \\
\midrule
Weather & 96 & \textbf{\textcolor{red}{0.155}} & \textbf{\textcolor{red}{0.203}} & \underline{\textcolor{blue}{0.157}} & \underline{\textcolor{blue}{0.205}} & 0.158 & \textbf{\textcolor{red}{0.203}} & 0.176 & 0.216 \\
 & 192 & \textbf{\textcolor{red}{0.202}} & \underline{\textcolor{blue}{0.248}} & \underline{\textcolor{blue}{0.205}} & 0.251 & 0.207 & \textbf{\textcolor{red}{0.247}} & 0.225 & 0.257 \\
 & 336 & \textbf{\textcolor{red}{0.255}} & \textbf{\textcolor{red}{0.290}} & \underline{\textcolor{blue}{0.258}} & \underline{\textcolor{blue}{0.291}} & 0.269 & 0.293 & 0.281 & 0.299 \\
 & 720 & \textbf{\textcolor{red}{0.327}} & \textbf{\textcolor{red}{0.341}} & \underline{\textcolor{blue}{0.334}} & \underline{\textcolor{blue}{0.344}} & 0.346 & 0.345 & 0.358 & 0.350 \\
 & \textit{Avg} & \textbf{\textcolor{red}{0.235}} & \textbf{\textcolor{red}{0.270}} & \underline{\textcolor{blue}{0.239}} & 0.273 & 0.245 & \underline{\textcolor{blue}{0.272}} & 0.260 & 0.280 \\
\midrule
\multicolumn{2}{l}{1st count} & \textbf{\textcolor{red}{39}} & \textbf{\textcolor{red}{23}} & 14 & 18 & 0 & 6 & 4 & 5 \\
\bottomrule
\end{tabular}
\caption{Per-horizon results with a fixed lookback of 96 ($H$: forecast horizon; \textit{Avg}: mean over the four horizons, as in the main table). Best in red bold, second underlined, per row over the columns shown. 1st count over individual horizons only.}
\label{tab:appendix_per_horizon_table}
\end{table}

\providecommand{\ourmethod}{GNA}
\begin{table*}[t]
\centering
\scriptsize
\setlength{\tabcolsep}{3pt}
\begin{tabular}{lcccccccccccc}
\toprule
 & \multicolumn{4}{c}{\textit{\ourmethod{} (this work)}} & \multicolumn{4}{c}{\textit{Backbones}} & \multicolumn{4}{c}{\textit{Retrieval-based}} \\
\cmidrule(lr){2-5}\cmidrule(lr){6-9}\cmidrule(lr){10-13}
Dataset & \multicolumn{2}{c}{w/ GridTST} & \multicolumn{2}{c}{w/ iTransformer} & \multicolumn{2}{c}{GridTST} & \multicolumn{2}{c}{iTransformer} & \multicolumn{2}{c}{GTR} & \multicolumn{2}{c}{RAFT$^{\ddagger}$} \\
 & \multicolumn{2}{c}{\tiny } & \multicolumn{2}{c}{\tiny } & \multicolumn{2}{c}{\tiny arXiv \citeyear{GridTST}, our run} & \multicolumn{2}{c}{\tiny ICLR \citeyear{liuITransformerInvertedTransformers2024}} & \multicolumn{2}{c}{\tiny ICLR \citeyear{cao2026enhancing}} & \multicolumn{2}{c}{\tiny ICML \citeyear{han2025retrieval}} \\
\cmidrule(lr){2-3}\cmidrule(lr){4-5}\cmidrule(lr){6-7}\cmidrule(lr){8-9}\cmidrule(lr){10-11}\cmidrule(lr){12-13}
 & MSE & MAE & MSE & MAE & MSE & MAE & MSE & MAE & MSE & MAE & MSE & MAE \\
\midrule
PEMS03 & \textbf{\textcolor{red}{0.082}} & \textbf{\textcolor{red}{0.182}} & \underline{\textcolor{blue}{0.084}} & \underline{\textcolor{blue}{0.183}} & 0.153 & 0.260 & 0.113 & 0.221 & 0.087 & 0.189 & 0.144 & 0.230 \\
PEMS04 & \textbf{\textcolor{red}{0.080}} & \underline{\textcolor{blue}{0.182}} & \underline{\textcolor{blue}{0.083}} & 0.184 & 0.174 & 0.283 & 0.111 & 0.221 & 0.087 & 0.189 & 0.104 & 0.210 \\
PEMS07 & \textbf{\textcolor{red}{0.065}} & \textbf{\textcolor{red}{0.155}} & \underline{\textcolor{blue}{0.067}} & \underline{\textcolor{blue}{0.156}} & 0.134 & 0.241 & 0.101 & 0.204 & 0.076 & 0.169 & 0.094 & 0.193 \\
PEMS08 & \textbf{\textcolor{red}{0.076}} & \textbf{\textcolor{red}{0.176}} & \textbf{\textcolor{red}{0.076}} & \textbf{\textcolor{red}{0.176}} & 0.184 & 0.275 & 0.150 & 0.226 & 0.142 & 0.222 & 0.151 & 0.234 \\
\midrule
ETTh1 & 0.463 & 0.459 & 0.473 & 0.464 & 0.456 & 0.448 & 0.454 & 0.447 & 0.439 & 0.434 & \underline{\textcolor{blue}{0.428}} & 0.433 \\
ETTh2 & 0.420 & 0.437 & 0.427 & 0.441 & 0.395 & 0.412 & 0.383 & 0.407 & 0.372 & 0.400 & 0.382 & 0.410 \\
ETTm1 & \textbf{\textcolor{red}{0.359}} & \underline{\textcolor{blue}{0.386}} & \underline{\textcolor{blue}{0.365}} & 0.389 & 0.386 & 0.402 & 0.407 & 0.410 & 0.367 & 0.389 & 0.381 & 0.400 \\
ETTm2 & 0.272 & 0.329 & 0.278 & 0.331 & 0.287 & 0.332 & 0.288 & 0.332 & 0.268 & 0.315 & 0.281 & 0.330 \\
ECL & \textbf{\textcolor{red}{0.153}} & \underline{\textcolor{blue}{0.252}} & \underline{\textcolor{blue}{0.154}} & 0.255 & 0.174 & 0.266 & 0.178 & 0.270 & 0.166 & 0.260 & 0.175 & 0.272 \\
Solar & \underline{\textcolor{blue}{0.198}} & 0.245 & 0.201 & \underline{\textcolor{blue}{0.243}} & 0.228 & 0.278 & 0.233 & 0.262 & \textbf{\textcolor{red}{0.194}} & 0.245 & 0.301 & 0.303 \\
Traffic & \textbf{\textcolor{red}{0.397}} & 0.269 & 0.409 & \textbf{\textcolor{red}{0.263}} & 0.412 & 0.271 & 0.428 & 0.282 & 0.470 & 0.280 & 0.414 & 0.284 \\
Weather & \textbf{\textcolor{red}{0.235}} & 0.270 & \underline{\textcolor{blue}{0.239}} & 0.273 & 0.245 & 0.272 & 0.258 & 0.278 & \underline{\textcolor{blue}{0.239}} & \underline{\textcolor{blue}{0.268}} & 0.270 & 0.309 \\
\midrule
1st count & \textbf{\textcolor{red}{8}} & 3 & 1 & 2 & 0 & 0 & 0 & 0 & 1 & 0 & 0 & 0 \\
\bottomrule
\end{tabular}
\par\medskip
\begin{tabular}{lcccccccccccccc}
\toprule
 & \multicolumn{14}{c}{\textit{Other forecasters}} \\
\cmidrule(lr){2-15}
Dataset & \multicolumn{2}{c}{TQNet} & \multicolumn{2}{c}{Amplifier} & \multicolumn{2}{c}{FilterTS} & \multicolumn{2}{c}{S-Mamba} & \multicolumn{2}{c}{CycleNet} & \multicolumn{2}{c}{SOFTS} & \multicolumn{2}{c}{TimeXer} \\
 & \multicolumn{2}{c}{\tiny ICML \citeyear{lin2025temporal}} & \multicolumn{2}{c}{\tiny AAAI \citeyear{10.1609/aaai.v39i11.33267}} & \multicolumn{2}{c}{\tiny AAAI \citeyear{10.1609/aaai.v39i20.35438}} & \multicolumn{2}{c}{\tiny Neurocomputing \citeyear{WANG2025129178}} & \multicolumn{2}{c}{\tiny NeurIPS \citeyear{lin2024cyclenet}} & \multicolumn{2}{c}{\tiny NeurIPS \citeyear{han2024softs}} & \multicolumn{2}{c}{\tiny NeurIPS \citeyear{NEURIPS2024_0113ef46}} \\
\cmidrule(lr){2-3}\cmidrule(lr){4-5}\cmidrule(lr){6-7}\cmidrule(lr){8-9}\cmidrule(lr){10-11}\cmidrule(lr){12-13}\cmidrule(lr){14-15}
 & MSE & MAE & MSE & MAE & MSE & MAE & MSE & MAE & MSE & MAE & MSE & MAE & MSE & MAE \\
\midrule
PEMS03 & 0.097 & 0.203 & 0.131$^{\ddagger}$ & 0.239$^{\ddagger}$ & 0.134$^{\ddagger}$ & 0.246$^{\ddagger}$ & 0.122 & 0.228 & 0.118$^{\ddagger}$ & 0.226$^{\ddagger}$ & 0.104 & 0.210 & 0.112$^{\ddagger}$ & 0.214$^{\ddagger}$ \\
PEMS04 & 0.091 & 0.197 & 0.135$^{\ddagger}$ & 0.249$^{\ddagger}$ & 0.125$^{\ddagger}$ & 0.241$^{\ddagger}$ & 0.103 & 0.211 & 0.119$^{\ddagger}$ & 0.232$^{\ddagger}$ & 0.102 & 0.208 & 0.105$^{\ddagger}$ & 0.209$^{\ddagger}$ \\
PEMS07 & 0.075 & 0.171 & 0.122$^{\ddagger}$ & 0.226$^{\ddagger}$ & 0.120$^{\ddagger}$ & 0.220$^{\ddagger}$ & 0.089 & 0.188 & 0.113$^{\ddagger}$ & 0.214$^{\ddagger}$ & 0.087 & 0.184 & 0.085$^{\ddagger}$ & 0.182$^{\ddagger}$ \\
PEMS08 & 0.142 & 0.229 & 0.183$^{\ddagger}$ & 0.271$^{\ddagger}$ & 0.180$^{\ddagger}$ & 0.266$^{\ddagger}$ & 0.148 & 0.224 & 0.150$^{\ddagger}$ & 0.246$^{\ddagger}$ & 0.138 & 0.219 & 0.175$^{\ddagger}$ & 0.250$^{\ddagger}$ \\
\midrule
ETTh1 & 0.441 & 0.434 & 0.430 & \underline{\textcolor{blue}{0.428}} & 0.433 & 0.430 & 0.455 & 0.450 & 0.457 & 0.441 & 0.449 & 0.442 & 0.437 & 0.437 \\
ETTh2 & 0.378 & 0.402 & \underline{\textcolor{blue}{0.359}} & \underline{\textcolor{blue}{0.391}} & 0.372 & 0.396 & 0.381 & 0.405 & 0.388 & 0.409 & 0.373 & 0.400 & 0.367 & 0.396 \\
ETTm1 & 0.377 & 0.393 & 0.381 & 0.394 & 0.385 & 0.396 & 0.398 & 0.405 & 0.379 & 0.396 & 0.393 & 0.403 & 0.382 & 0.397 \\
ETTm2 & 0.277 & 0.323 & 0.276 & 0.323 & 0.276 & 0.321 & 0.288 & 0.332 & \underline{\textcolor{blue}{0.266}} & \underline{\textcolor{blue}{0.314}} & 0.287 & 0.330 & 0.274 & 0.322 \\
ECL & 0.164 & 0.259 & 0.171 & 0.265 & 0.180 & 0.271 & 0.170 & 0.265 & 0.168 & 0.259 & 0.174 & 0.264 & 0.171 & 0.270 \\
Solar & \underline{\textcolor{blue}{0.198}} & 0.256 & 0.241$^{\ddagger}$ & 0.270$^{\ddagger}$ & 0.215$^{\ddagger}$ & 0.277$^{\ddagger}$ & 0.240 & 0.273 & 0.210 & 0.261 & 0.229 & 0.256 & 0.237$^{\ddagger}$ & 0.302$^{\ddagger}$ \\
Traffic & 0.445 & 0.276 & 0.482 & 0.315 & 0.471 & 0.315 & 0.414 & 0.276 & 0.472 & 0.301 & 0.409 & \underline{\textcolor{blue}{0.267}} & 0.466 & 0.287 \\
Weather & 0.242 & 0.269 & 0.243 & 0.271 & 0.244 & 0.274 & 0.251 & 0.276 & 0.243 & 0.271 & 0.255 & 0.278 & 0.241 & 0.271 \\
\midrule
1st count & 0 & 0 & 0 & 0 & 0 & 0 & 0 & 0 & 0 & 0 & 0 & 0 & 0 & 0 \\
\bottomrule
\end{tabular}
\par\medskip
\begin{tabular}{lcccccccccccccc}
\toprule
 & \multicolumn{12}{c}{\textit{Other forecasters (cont.)}} & \multicolumn{2}{c}{\textit{Concurrent preprints}} \\
\cmidrule(lr){2-13}\cmidrule(lr){14-15}
Dataset & \multicolumn{2}{c}{TimeMixer} & \multicolumn{2}{c}{PatchTST$^{\ddagger}$} & \multicolumn{2}{c}{TimesNet} & \multicolumn{2}{c}{Crossformer$^{\ddagger}$} & \multicolumn{2}{c}{DLinear$^{\ddagger}$} & \multicolumn{2}{c}{Ister} & \multicolumn{2}{c}{PAMNet$^{\dagger}$} \\
 & \multicolumn{2}{c}{\tiny ICLR \citeyear{wang2024timemixer}} & \multicolumn{2}{c}{\tiny ICLR \citeyear{nie2023a}} & \multicolumn{2}{c}{\tiny ICLR \citeyear{wu2023timesnet}} & \multicolumn{2}{c}{\tiny ICLR \citeyear{zhang2023crossformer}} & \multicolumn{2}{c}{\tiny AAAI \citeyear{10.1609/aaai.v37i9.26317}} & \multicolumn{2}{c}{\tiny ICASSP \citeyear{Cao_2026}} & \multicolumn{2}{c}{\tiny arXiv \citeyear{zhou2026pamnetcycleawarephaseamplitudemodulation} (preprint)} \\
\cmidrule(lr){2-3}\cmidrule(lr){4-5}\cmidrule(lr){6-7}\cmidrule(lr){8-9}\cmidrule(lr){10-11}\cmidrule(lr){12-13}\cmidrule(lr){14-15}
 & MSE & MAE & MSE & MAE & MSE & MAE & MSE & MAE & MSE & MAE & MSE & MAE & MSE & MAE \\
\midrule
PEMS03 & 0.167$^{\ddagger}$ & 0.267$^{\ddagger}$ & 0.180 & 0.291 & 0.147$^{\ddagger}$ & 0.248$^{\ddagger}$ & 0.169 & 0.281 & 0.278 & 0.375 & 0.108 & 0.217 & 0.091 & 0.188 \\
PEMS04 & 0.185$^{\ddagger}$ & 0.287$^{\ddagger}$ & 0.195 & 0.307 & 0.129$^{\ddagger}$ & 0.241$^{\ddagger}$ & 0.209 & 0.314 & 0.295 & 0.388 & 0.106 & 0.213 & 0.085 & \textbf{\textcolor{red}{0.180}} \\
PEMS07 & 0.181$^{\ddagger}$ & 0.271$^{\ddagger}$ & 0.211 & 0.303 & 0.124$^{\ddagger}$ & 0.225$^{\ddagger}$ & 0.235 & 0.315 & 0.329 & 0.395 & 0.092 & 0.193 & 0.078 & 0.163 \\
PEMS08 & 0.226$^{\ddagger}$ & 0.299$^{\ddagger}$ & 0.280 & 0.321 & 0.193$^{\ddagger}$ & 0.271$^{\ddagger}$ & 0.268 & 0.307 & 0.379 & 0.416 & \underline{\textcolor{blue}{0.136}} & 0.226 & 0.137 & \underline{\textcolor{blue}{0.211}} \\
\midrule
ETTh1 & 0.447 & 0.440 & 0.469 & 0.454 & 0.458 & 0.450 & 0.529 & 0.522 & 0.456 & 0.452 & 0.438 & 0.438 & \textbf{\textcolor{red}{0.420}} & \textbf{\textcolor{red}{0.424}} \\
ETTh2 & 0.364 & 0.395 & 0.387 & 0.407 & 0.414 & 0.427 & 0.942 & 0.684 & 0.559 & 0.515 & \textbf{\textcolor{red}{0.349}} & \textbf{\textcolor{red}{0.387}} & 0.374 & 0.393 \\
ETTm1 & 0.381 & 0.395 & 0.387 & 0.400 & 0.400 & 0.406 & 0.513 & 0.496 & 0.403 & 0.407 & 0.386 & 0.399 & \underline{\textcolor{blue}{0.365}} & \textbf{\textcolor{red}{0.384}} \\
ETTm2 & 0.275 & 0.323 & 0.281 & 0.326 & 0.291 & 0.333 & 0.757 & 0.610 & 0.350 & 0.401 & 0.279 & 0.325 & \textbf{\textcolor{red}{0.264}} & \textbf{\textcolor{red}{0.308}} \\
ECL & 0.182 & 0.272 & 0.205 & 0.290 & 0.192 & 0.295 & 0.244 & 0.334 & 0.212 & 0.300 & 0.167 & 0.260 & 0.161 & \textbf{\textcolor{red}{0.251}} \\
Solar & 0.216 & 0.280 & 0.270 & 0.307 & 0.301$^{\ddagger}$ & 0.319$^{\ddagger}$ & 0.641 & 0.639 & 0.330 & 0.401 & \textemdash & \textemdash & 0.204 & \textbf{\textcolor{red}{0.228}} \\
Traffic & 0.484 & 0.297 & 0.481 & 0.304 & 0.620 & 0.336 & 0.550 & 0.304 & 0.625 & 0.383 & \underline{\textcolor{blue}{0.399}} & 0.270 & 0.440 & \textbf{\textcolor{red}{0.263}} \\
Weather & 0.240 & 0.271 & 0.259 & 0.281 & 0.259 & 0.287 & 0.259 & 0.315 & 0.265 & 0.317 & 0.243 & 0.271 & 0.240 & \textbf{\textcolor{red}{0.263}} \\
\midrule
1st count & 0 & 0 & 0 & 0 & 0 & 0 & 0 & 0 & 0 & 0 & 1 & 1 & 2 & \textbf{\textcolor{red}{8}} \\
\bottomrule
\end{tabular}
\caption{All compared models, lookback 96, horizon-averaged. Baseline numbers are taken from the respective publications (GridTST: our own lookback-96 run). $^{\dagger}$Concurrent preprint, not peer-reviewed. Best in red bold, second underlined, over all models shown. $^{\ddagger}$Lookback-96 result reported by another paper, because the method's own paper uses a different lookback or omits the dataset: RAFT by GTR \citep{cao2026enhancing}; PatchTST, DLinear, Crossformer, and TimesNet on Solar/PEMS by iTransformer \citep{liuITransformerInvertedTransformers2024}; CycleNet on PEMS and TimeXer on Solar/PEMS by TQNet \citep{lin2025temporal}; Amplifier and FilterTS on Solar/PEMS and TimeMixer on PEMS by PAMNet \citep{zhou2026pamnetcycleawarephaseamplitudemodulation}. CycleNet: CycleNet/MLP variant.}
\label{tab:appendix_all_models_table}
\end{table*}

\begin{figure}[!htbp]
\centering
\includegraphics{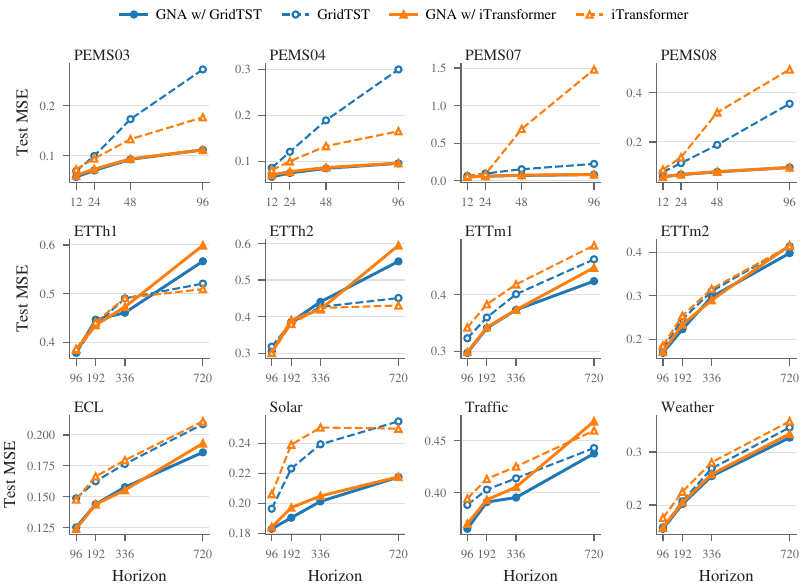}
\caption{%
\cd{Test MSE per horizon $H$ for every dataset: \ourmethod{} and the plain backbones.}}
\label{fig:mse-by-horizon}
\end{figure}

\section{Seed robustness}
\label{app:seeds}
\begin{cdraft}
We repeat all twelve datasets with seeds 2023 (the tracked runs in all other tables), 2024 and 2025, for \ourmethod{} and for the plain backbone (Table~\ref{tab:seeds}). \ourmethod{} with GridTST beats the plain backbone in every seed on 10 of 12 datasets and is worse in every seed on ETTh1 and ETTh2. Retrieval also stabilizes training. The seed-to-seed variation of \ourmethod{} is a median 0.3\% of the MSE (at most 0.8\%), against a median 4\% and up to 20\% on PEMS for the plain GridTST. On PEMS, the seed-2023 backbone run used in the main table is the best of its three seeds, so the main-table comparison there is conservative. Relative to seed noise, all but four ablation effects in Fig.~\ref{fig:ablation-dots} exceed twice the full model's seed standard deviation; the four within this noise (hollow markers) are Weather without global slots, PEMS08 without rank fusion, and ETTm1 and Weather without the future-aware embedding.
\end{cdraft}

\begin{table}[!htbp]
\centering
\small
\setlength{\tabcolsep}{4pt}
\begin{tabular}{@{}lcccrc@{}}
\toprule
Dataset & GridTST & GNA w/ GridTST & GNA w/ iTransformer & \multicolumn{1}{c}{Reduction} & Seeds won \\
\midrule
PEMS03 & 0.191 $\pm$ 0.0331 & 0.082 $\pm$ 0.0001 & 0.084 $\pm$ 0.0001 & $+56.9$\% & 3/3 \\
PEMS04 & 0.225 $\pm$ 0.0445 & 0.080 $\pm$ 0.0001 & 0.082 $\pm$ 0.0001 & $+64.4$\% & 3/3 \\
PEMS07 & 0.166 $\pm$ 0.0281 & 0.065 $\pm$ 0.0000 & 0.067 $\pm$ 0.0003 & $+60.7$\% & 3/3 \\
PEMS08 & 0.231 $\pm$ 0.0413 & 0.076 $\pm$ 0.0002 & 0.076 $\pm$ 0.0003 & $+67.3$\% & 3/3 \\
\midrule
ETTh1 & 0.455 $\pm$ 0.0028 & 0.460 $\pm$ 0.0027 & 0.471 $\pm$ 0.0039 & $-1.1$\% & 0/3 \\
ETTh2 & 0.389 $\pm$ 0.0054 & 0.417 $\pm$ 0.0034 & 0.423 $\pm$ 0.0085 & $-7.2$\% & 0/3 \\
ETTm1 & 0.389 $\pm$ 0.0026 & 0.361 $\pm$ 0.0022 & 0.368 $\pm$ 0.0026 & $+7.2$\% & 3/3 \\
ETTm2 & 0.286 $\pm$ 0.0010 & 0.273 $\pm$ 0.0022 & 0.280 $\pm$ 0.0028 & $+4.7$\% & 3/3 \\
ECL & 0.189 $\pm$ 0.0127 & 0.153 $\pm$ 0.0001 & 0.153 $\pm$ 0.0008 & $+18.9$\% & 3/3 \\
Solar & 0.251 $\pm$ 0.0197 & 0.198 $\pm$ 0.0006 & 0.201 $\pm$ 0.0005 & $+21.0$\% & 3/3 \\
Traffic & 0.414 $\pm$ 0.0020 & 0.397 $\pm$ 0.0007 & 0.410 $\pm$ 0.0005 & $+4.1$\% & 3/3 \\
Weather & 0.248 $\pm$ 0.0027 & 0.236 $\pm$ 0.0012 & 0.239 $\pm$ 0.0003 & $+5.0$\% & 3/3 \\
\bottomrule
\end{tabular}
\caption{\papercaption{tab:seeds}}
\label{tab:seeds}
\end{table}

\section{Ablation details and qualitative example}
\label{app:ablations}
\begin{figure}[!htbp]
\centering
\includegraphics{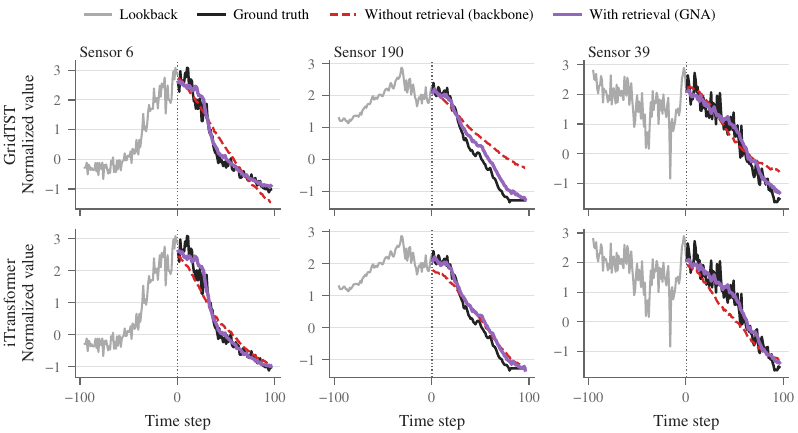}
\caption{%
\cd{A typical PEMS04 test window ($H=96$): ground truth and the forecasts of \ourmethod{} and of the plain backbone, for both backbones. The window and variate were chosen by a fixed rule (median gain of \ourmethod{} over the backbone; variate with the largest target spread), not by appearance.}}
\label{fig:qualitative}
\end{figure}

\begin{cdraft}
All ablation arms use the full \ourmethod{} recipe with GridTST (seed 2023, $L=96$, 10 epochs) and change exactly one thing; the control is the tracked full model. Each arm and what it tests:
\begin{itemize}[leftmargin=*,itemsep=1pt,topsep=2pt]
\item \emph{w/o relevance (mismatched)}: each query keeps its own window but receives the complete neighbor set (slots, slot types, masks, calendar match) of a different query. At validation and test time, that query is drawn uniformly from the training split, so all neighbors stay strictly past; during training, it lies at least $L+H$ windows away. Architecture, parameter count and training recipe are unchanged, and the swap is applied identically in training, validation and test, so only the relevance of the neighbors changes.
\item \emph{w/o retrieval}: the plain backbone.
\item \emph{w/o per-variate slots}: ten global slots, i.e.\ the global half of the full model doubled.
\item \emph{w/o global slots}: ten per-variate slots from the spaced pool.
\item \emph{w/o future-aware embedding}: the embedding ranking is replaced by the cosine similarity of the standardized raw lookback windows; fusion, eligibility, spacing and assembly are unchanged.
\item \emph{w/o rank fusion}: ranking by the learned embedding alone, without regime, seasonal and calendar terms.
\end{itemize}
All ablation banks were audited for causality. Table~\ref{tab:ablations} lists the numbers. Fig.~\ref{fig:qualitative} shows a typical forecast.

\paragraph{Retrieved-future quality.} The ablations above measure forecasts. Table~\ref{tab:retrieved-future-quality} measures the retrieved futures themselves, before any model sees them, by comparing each slot's future with the query's true future. The learned ranking returns slot futures whose average is 0.1--4.0\% closer to the truth than with raw-window ranking; the best single slot changes little. Per-variate slots contain a much closer candidate for every variate: the best per-variate slot is on average 17--45\% closer than the best global slot, and closer on every dataset and at every horizon (12--48\%). Their plain average is not always better (PEMS08: $+11\%$), so the gain depends on choosing among them, which is the role of the gate. \begin{table}[!htbp]
\centering
\small
\begin{tabular}{@{}lrrrr@{}}
\toprule
 & \multicolumn{2}{c}{Learned vs.\ raw-window ranking} & \multicolumn{2}{c}{Per-variate vs.\ global slots} \\
\cmidrule(lr){2-3}\cmidrule(lr){4-5}
Dataset & Mean of slots & Best slot & Mean of slots & Best slot \\
\midrule
PEMS04 & $-4.0$ & $-0.2$ & $-4.4$ & $-17.2$ \\
PEMS08 & $-1.3$ & $+0.7$ & $+11.0$ & $-24.2$ \\
ETTm1 & $-0.1$ & $+0.7$ & $-15.8$ & $-24.4$ \\
ETTm2 & $-3.4$ & $-0.7$ & $-13.8$ & $-26.7$ \\
Weather & $-2.6$ & $-1.4$ & $-25.8$ & $-45.2$ \\
\bottomrule
\end{tabular}
\caption{\papercaption{tab:retrieved-future-quality}}
\label{tab:retrieved-future-quality}
\end{table}

\end{cdraft}

\begin{table}[!htbp]
\centering
\small
\begin{tabular}{@{}lrrrrr@{}}
\toprule
 & \multicolumn{1}{c}{PEMS04} & \multicolumn{1}{c}{PEMS08} & \multicolumn{1}{c}{ETTm1} & \multicolumn{1}{c}{ETTm2} & \multicolumn{1}{c}{Weather} \\
\midrule
Full model, MSE & 0.080 & 0.076 & 0.359 & 0.272 & 0.235 \\
Full model, MAE & 0.182 & 0.176 & 0.386 & 0.329 & 0.270 \\
\midrule
\multicolumn{6}{@{}l}{\textit{MSE increase vs.\ full model (\%)}} \\
w/o relevance (mismatched neighbors) & +201.5 & +246.8 & +10.3 & +21.1 & +5.9 \\
w/o per-variate slots & +3.5 & +8.3 & +4.8 & +1.9 & +5.5 \\
w/o global slots & +13.0 & +15.8 & +2.5 & +3.4 & +0.3 \\
w/o future-aware embedding & +7.4 & +5.1 & +1.1 & +3.9 & +0.9 \\
w/o rank fusion (embedding only) & -1.1 & +0.4 & +4.1 & +12.0 & +4.5 \\
w/o retrieval (backbone only) & +116.5 & +142.1 & +7.7 & +5.7 & +4.4 \\
\midrule
\multicolumn{6}{@{}l}{\textit{MAE increase vs.\ full model (\%)}} \\
w/o relevance (mismatched neighbors) & +86.4 & +96.0 & +6.4 & +11.1 & +5.9 \\
w/o per-variate slots & +2.5 & +5.0 & +2.6 & +0.8 & +3.0 \\
w/o global slots & +7.3 & +8.4 & +1.9 & +1.2 & +0.5 \\
w/o future-aware embedding & +3.8 & +0.9 & +0.3 & +0.6 & +0.1 \\
w/o rank fusion (embedding only) & -0.7 & +1.4 & +2.6 & +5.6 & +3.1 \\
w/o retrieval (backbone only) & +55.8 & +56.3 & +4.2 & +0.8 & +0.6 \\
\bottomrule
\end{tabular}
\caption{\papercaption{tab:ablations}}
\label{tab:ablations}
\end{table}

\begin{table}[!htbp]
\centering
\small
\begin{tabular}{@{}lccc@{\hspace{2.2em}}lccc@{}}
\toprule
\multicolumn{4}{@{}l}{Slots per query $K$ (\ourmethod{})} & \multicolumn{4}{l@{}}{Lookback $L$ (ETTm1)} \\
\cmidrule(r{2.2em}){1-4}\cmidrule{5-8}
Dataset & $K=5$ & $K=10$ & $K=20$ & $L$ & GridTST & \ourmethod{} & Change \\
\midrule
PEMS08 & 0.101 & 0.097 & 0.096 & 48 & 0.466 & 0.310 & $-33.4\%$ \\
ETTm1 & 0.303 & 0.297 & 0.295 & 96 & 0.323 & 0.297 & $-8.0\%$ \\
 & & & & 192 & 0.301 & 0.297 & $-1.2\%$ \\
 & & & & 336 & 0.288 & 0.303 & $+5.3\%$ \\
\bottomrule
\end{tabular}
\caption{\cd{Sensitivity at $H=96$ (test MSE, GridTST backbone, seed 2023). Left: number of slots per query, half global and half per-variate ($K=10$: reported runs). Right: lookback $L$ for the plain GridTST and \ourmethod{}, each with its own embedder and bank per $L$; change: \ourmethod{} relative to GridTST.}}
\label{tab:sensitivity}
\end{table}

\section{Computational cost}

\label{app:compute}
\begin{cdraft}
All experiments run on one NVIDIA RTX PRO 6000 Blackwell GPU (96\,GB) with an AMD Ryzen Threadripper PRO 7955WX CPU (16 cores) and 124\,GB RAM, using Python 3.13, PyTorch 2.12 (CUDA 13.0) and NumPy 2.4. Table~\ref{tab:compute} reports resource use at $L=H=96$, with one stage at a time on an otherwise idle GPU. The bank is built once before training; the index holds what retrieval needs (embeddings, rankings, regime features). Our implementation also caches every query's assembled slots for fast training. This cache is 14 to 90 times larger than the index and could be replaced by assembling the slots on the fly. Table~\ref{tab:inference-stages} breaks down the latency of one online forecast. Retrieval takes 19--25\,ms per query on the CPU and is dominated by the exact search, which approximate search would make sublinear in $N$.
\end{cdraft}

\begin{table}[!htbp]
\centering
{\small\setlength{\tabcolsep}{2.8pt}\begin{tabular}{@{}l@{\hspace{6pt}}r@{\hspace{8pt}}cccc@{\hspace{8pt}}c@{\hspace{4pt}}c@{\hspace{8pt}}c@{\hspace{4pt}}c@{\hspace{8pt}}c@{\hspace{4pt}}c@{}}
\toprule
 & & \multicolumn{4}{c}{Retrieval bank (built once)} & \multicolumn{2}{c}{Train} & \multicolumn{2}{c}{GPU energy} & \multicolumn{2}{c}{GPU mem.} \\
\cmidrule(lr){3-6}\cmidrule(lr){7-8}\cmidrule(lr){9-10}\cmidrule(lr){11-12}
 & & wall & CPU & index & cache & \multicolumn{2}{c}{(s/epoch)} & \multicolumn{2}{c}{(Wh/run)} & \multicolumn{2}{c}{(GB)} \\
Dataset & $D$ & (min) & (min) & (GB) & (GB) & w/o & with & w/o & with & w/o & with \\
\midrule
ETTm1 & 7 & 5.2 & 12.4 & 0.15 & 2.0 & 10 & 15 & 3 & 5 & 0.8 & 0.8 \\
Weather & 21 & 5.5 & 12.2 & 0.19 & 5.0 & 13 & 19 & 6 & 9 & 1.0 & 1.1 \\
PEMS08 & 170 & 3.3 & 5.2 & 0.17 & 12.8 & 14 & 17 & 10 & 14 & 4.1 & 4.8 \\
ECL & 321 & 7.0 & 10.4 & 0.40 & 36.3 & 78 & 89 & 98 & 111 & 15.1 & 19.2 \\
\bottomrule
\end{tabular}
}
\caption{%
\cd{Resource use at $L=96$ (GridTST backbone, batch size 32) on an otherwise idle GPU: bank construction (wall-clock and CPU time, index and training-cache size), training time per epoch, GPU energy per training run and GPU memory. w/o: the same model with retrieval disabled.}}
\label{tab:compute}
\end{table}

\begin{table}[!htbp]
\centering
{\small\begin{tabular}{@{}lrr@{\hspace{10pt}}rrrr@{\hspace{10pt}}rr@{\hspace{10pt}}r@{}}
\toprule
 & & & \multicolumn{4}{c}{Retrieval stages (ms)} & \multicolumn{2}{c}{Retrieval (ms)} & Forecast \\
\cmidrule(lr){4-7}\cmidrule(lr){8-9}
Dataset & $D$ & $N$ & Embed & Search & Fusion & Assembly & median & p90 & (ms)$^{\dagger}$ \\
\midrule
ETTm1 & 7 & 57,601 & 0.7 & 18.9 & 0.6 & 0.9 & 21.2 & 26.7 & $\leq$0.6 \\
Weather & 21 & 52,697 & 0.7 & 22.5 & 0.6 & 1.2 & 25.2 & 41.3 & $\leq$0.7 \\
PEMS08 & 170 & 17,665 & 2.4 & 11.7 & 0.4 & 4.4 & 19.1 & 23.1 & $\leq$2.1 \\
ECL & 321 & 26,305 & 2.2 & 12.3 & 0.6 & 8.0 & 23.3 & 29.0 & $\leq$3.2 \\
\bottomrule
\end{tabular}
}
\caption{%
\cd{Latency of one forecast by stage. Retrieval runs on the CPU for one query at a time (8 threads; median and 90th percentile over 200 test queries after 20 warm-up queries): embedding the query with $\phi$, exact search over the eligible bank, rank fusion, and slot assembly. $D$: variates; $N$: bank windows. $^{\dagger}$Full test-set evaluation time per window on the GPU (batch size 32, including data loading), an upper bound on the forward pass.}}
\label{tab:inference-stages}
\end{table}

\section{Comparison with retrieval-based forecasting methods}
\label{app:related-comparison}
\begin{cdraft}
Table~\ref{tab:method-comparison} compares \ourmethod{} with retrieval-based forecasting methods, based on each paper's description. Among these methods, only \ourmethod{} trains its retrieval embedding to predict each window's own future and combines whole-window and per-variate neighbors in one multivariate model; its bank also grows causally during the test period, whereas RAFT and RATD search the training data and TS-RAG and TimeRAF a knowledge base fixed before testing. TimeRAF also learns its retriever, but from the forecaster's error on retrieved candidates; TS-RAG, RAF and TimeRAF retrieve univariate segments, i.e.\ for each variate on its own.
\end{cdraft}

\begin{table}[!htbp]
\centering
\scriptsize
\setlength{\tabcolsep}{3pt}
\renewcommand{\arraystretch}{1.15}
\begin{tabular}{@{}p{1.35cm}p{2.15cm}p{2.35cm}p{2.0cm}p{2.35cm}p{1.9cm}@{}}
\toprule
Method & Retrieved & Similarity & Assembly & How it enters the model & Bank at test time \\
\midrule
RAFT \citep{han2025retrieval} & (history, future) patches of the series & Pearson correlation of offset-removed raw windows, at several downsampling periods & top-$m$ windows, softmax-weighted sum & projected, combined with an MLP forecaster & training data \\
GTR \citep{cao2026enhancing} & segment of a learned global-cycle embedding (no past windows) & absolute position within a fixed global cycle & one aligned segment & fused with the input (linear map, 2D convolution, residual), then an MLP backbone & learned parameters \\
RATD \citep{NEURIPS2024_053ee34c} & (history, future) windows & distance between pre-trained encoder embeddings of the histories & $k$ nearest whole windows & guide a diffusion denoiser (reference-modulated attention) & training data \\
TS-RAG \citep{ning2025tsrag} & (context, future) pairs, univariate & Euclidean distance of frozen foundation-model encoder embeddings & top-$k$ futures & adaptive retrieval mixer on a frozen foundation model & fixed knowledge base (multi-domain, or the dataset's training set) \\
TimeRAF \citep{zhang2024timeraf} & univariate segments of input length (multi-domain knowledge base) & learned retriever, trained so that candidates that lower the foundation model's forecast error score higher & top-8, averaged & channel prompting (MLP, residual) on a foundation model & fixed knowledge base (multi-domain, or the dataset's training set) \\
RAF \citep{tireRetrievalAugmentedTime2026} & (context, future) segments, univariate & $\ell_2$ distance of foundation-model encoder embeddings & best match & prepended to the input context of a foundation model & other series of the dataset, before the query's forecast window \\
\midrule
GNA (this work) & (history, future) windows & learned embedding trained to predict the future, fused with regime, seasonal and calendar ranks & global slots + per-variate re-ranked slots & learned gate over persistence and futures, backbone term, cross-attention & strictly causal, grows with the test period \\
\bottomrule
\end{tabular}

\caption{%
\cd{Retrieval-based forecasting methods compared by what they retrieve, how they measure similarity, how they assemble and use the retrieved information, and which data their bank contains at test time.}}
\label{tab:method-comparison}
\end{table}

\section{Oracle-ceiling details}
\label{app:oracle}
\begin{figure}[!htbp]
\centering
\includegraphics{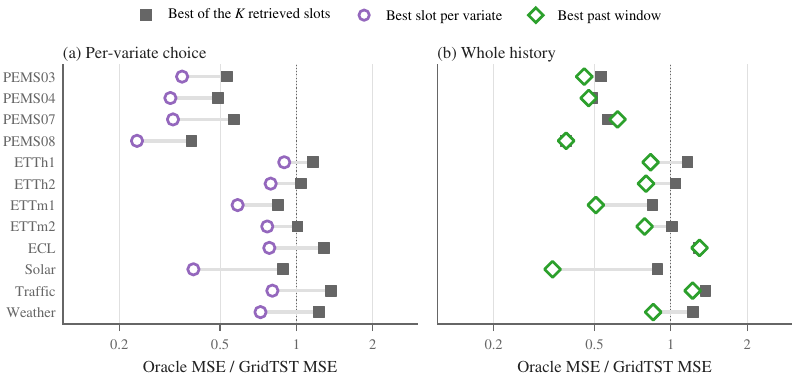}
\caption{%
\cd{Retrieval headroom at $H=96$, relative to GridTST, with oracle choices that use the true future: (a)~best of the $K$ retrieved slots vs.\ the best slot chosen per variate; (b)~vs.\ the best single past window.}}
\label{fig:retrieval-headroom}
\end{figure}

\begin{figure}[!htbp]
\centering
\includegraphics{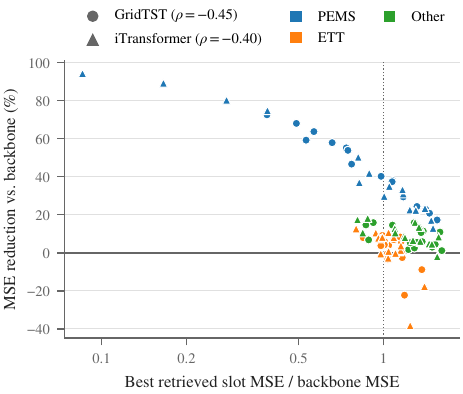}
\caption{%
\cd{Retrieval quality vs.\ forecast gain, one point per dataset, horizon and backbone; $\rho$: Spearman correlation.}}
\label{fig:retrieval-quality-vs-gain}
\end{figure}

\begin{figure}[!htbp]
\centering
\includegraphics{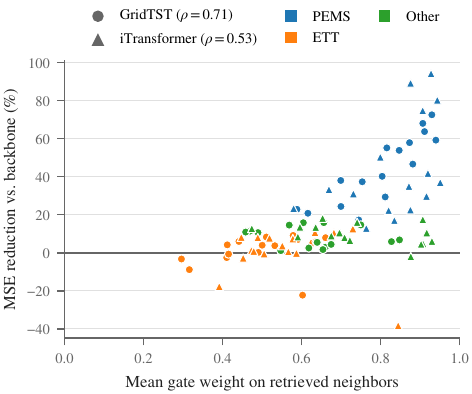}
\caption{%
\cd{Learned trust vs.\ forecast gain, one point per dataset, horizon and backbone. $x$: mean gate weight on retrieved slots (the rest goes to persistence $\mathbf{c}_0$; the backbone term is added outside the gate); $y$: MSE reduction relative to the plain backbone; $\rho$: Spearman correlation.}}
\label{fig:gate-weight-vs-gain}
\end{figure}

\begin{figure}[!htbp]
\centering
\includegraphics{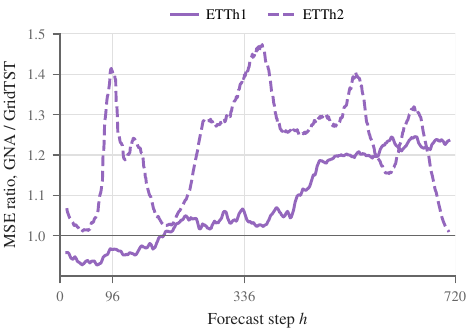}
\caption{%
\cd{ETTh at $H=720$: ratio of the test MSE of \ourmethod{} to that of GridTST per forecast step (24-step moving average). Below 1: retrieval helps.}}
\label{fig:etth-horizon-loss}
\end{figure}

\begin{cdraft}
To separate the quality of the retrieved candidates from the model's ability to use them, we use oracles that choose past windows with hindsight, using the true future, from the same causally eligible bank. Choosing among thousands of windows with the target includes selection luck, so absolute oracle numbers are optimistic. The comparison between raw and level-free futures is fair, since both use the same windows. Figure~\ref{fig:retrieval-headroom} shows the headroom between the retrieved slots and the best past windows. Figures~\ref{fig:retrieval-quality-vs-gain} and~\ref{fig:gate-weight-vs-gain} relate retrieval quality and gate trust to the forecast gain across all datasets, horizons and backbones. For ETTh (Table~\ref{tab:oracle-ceiling}, Fig.~\ref{fig:etth-horizon-loss}), the analysis below uses full test-set predictions of the tracked runs.

\paragraph{ETTh diagnosis.} On ETTh1 and ETTh2, \ourmethod{} is worse than GridTST. The loss is a far-horizon effect (Fig.~\ref{fig:etth-horizon-loss}). At $H=720$ on ETTh1, \ourmethod{} is 5.5\% better than GridTST over steps 1--96 and 16.5\% worse over steps 337--720, which carry all of the excess error. ETTh2 is worse at nearly every step, with 60\% of the gate weight on neighbors (32\% on ETTh1). The loss grows with the hindsight level offset between the gate-weighted neighbor futures and the truth (Spearman 0.25 and 0.32), but not with the level gap at the forecast origin, which the gate can observe: retrieved futures look aligned at the start and drift further out. A hindsight oracle over the same eligible windows supports this reading (Table~\ref{tab:oracle-ceiling}): the mean of the ten best raw past futures barely beats GridTST at $H=720$ ($-5.8\%$, $-0.9\%$), whereas the same windows with level-free futures, shifted to start at the query's last value, would reduce the error by 26\% and 36\%. A fixed shift hurts ECL and Traffic, so a learned level correction is the natural next step.
\end{cdraft}

\begin{table}[!htbp]
\centering
\small
\begin{tabular}{@{}lcc@{}}
\toprule
Test MSE & ETTh1/720 & ETTh2/720 \\
\midrule
GridTST (no retrieval) & 0.521 & 0.451 \\
GNA (this work) & 0.566 & 0.551 \\
Best of the $K$ retrieved slots & 0.712 & 0.535 \\
\midrule
\multicolumn{3}{@{}l}{\textit{Best past window(s), raw futures}} \\
Best single & 0.640 & 0.519 \\
Mean of 10 best & 0.491 & 0.447 \\
\midrule
\multicolumn{3}{@{}l}{\textit{Best past window(s), level-free futures}} \\
Best single & 0.596 & 0.380 \\
Mean of 10 best & 0.386 & 0.289 \\
\bottomrule
\end{tabular}
\caption{\papercaption{tab:oracle-ceiling}}
\label{tab:oracle-ceiling}
\end{table}

\end{document}